\documentclass[final,1p,times]{elsarticle}

\usepackage{subfig} 
\usepackage{graphicx}
\usepackage{float}

\usepackage{multirow}  
\usepackage{booktabs} 

\usepackage{amssymb}
\usepackage{amsthm}
\usepackage{bm}

\usepackage{amsmath}

\usepackage{lineno}

\usepackage{color,soul} 
\definecolor{Dred}{RGB}{192,0,0} 
\definecolor{Blue}{RGB}{0,0,0} 

\usepackage[colorlinks,
            linkcolor=blue,
            anchorcolor=blue,
            citecolor=blue]{hyperref}

\usepackage{algorithm}
\usepackage{algpseudocode}

\journal{Information Fusion}

\begin{document}

\begin{frontmatter}




\cortext[cor1]{Corresponding author.}

\title{Multi-View Class Incremental Learning}

\author[addr1,addr2,addr3]{Depeng Li} 
\ead{dpli@hust.edu.cn} 
\author[addr1]{Tianqi Wang}
\ead{tianqiwang@hust.edu.cn}
\author[addr1]{Junwei Chen}
\ead{junwei_chen@hust.edu.cn}
\author[addr3]{Kenji Kawaguchi}
\ead{kawaguch@csail.mit.edu}
\author[addr4]{Cheng Lian}
\ead{chenglian@whut.edu.cn}
\author[addr1,addr2]{Zhigang Zeng\corref{cor1}}
\ead{zgzeng@hust.edu.cn}

\affiliation[addr1]{organization={School of Artificial Intelligence and Automation},
            addressline={Huazhong University of Science and Technology}, 
            city={Wuhan},
            postcode={430074}, 
            country={China}}

\affiliation[addr2]{organization={Institute of Artificial Intelligence},
            addressline={Huazhong University of Science and Technology}, 
            city={Wuhan},
            postcode={430074}, 
            country={China}}

\affiliation[addr3]{organization={School of Computing},
            addressline={National University of Singapore}, 
            postcode={117417}, 
            country={Singapore}}

\affiliation[addr4]{organization={School of Automation},
            addressline={Wuhan University of Technology}, 
            city={Wuhan},
            postcode={430070}, 
            country={China}}



\begin{abstract}
Multi-view learning (MVL) has gained great success in integrating information from multiple perspectives of a dataset to improve downstream task performance. To make MVL methods more practical in an open-ended environment, this paper investigates a novel paradigm called multi-view class incremental learning (MVCIL), where a single model incrementally classifies new classes from a continual stream of views, requiring no access to earlier views of data. However, MVCIL is challenged by the catastrophic forgetting of old information and the interference with learning new concepts. To address this, we first develop a randomization-based representation learning technique serving for feature extraction to guarantee their separate view-optimal working states, during which multiple views belonging to a class are presented sequentially; Then, we integrate them one by one in the orthogonality fusion subspace spanned by the extracted features; Finally, we introduce selective weight consolidation for learning-without-forgetting decision-making while encountering new classes. Extensive experiments on synthetic and real-world datasets validate the effectiveness of our approach.
\end{abstract}




\end{frontmatter}


\section{Introduction}
\label{Sec_Intro}
Benefiting from the advancements in information technology, multi-view data have witnessed a widespread increase over the last few decades \cite{zhao2017multi, zhang2019feature}. For instance, in computer vision and image processing applications, video recognition represented by multiple frames, car photos taken from different angles, and face images captured under different features are common in real-world scenarios \cite{cao2017generalized, li2021asymmetric}. 
Although each of the views by itself might be sufficient for a certain learning task, improvements can be obtained by efficiently fusing complementary and consensus information among them, yielding numerous multi-view learning (MVL) methods \cite{tang2018consensus, tang2018learning, tang2019cross, houthuys2021tensor, liu2021one, yan2023collaborative}. They typically focus on integrating information for follow-up tasks and have achieved tremendous success across a wide range of applications \cite{shi2020multi, fu2021red, tian2022classification, yang2023multi}. 

Most existing MVL algorithms, however, explicitly assume that all data views are static and can be simultaneously accessed, which has some intrinsic drawbacks that need to be addressed \cite{xu2016streaming, zhou2019incremental}. First, this ideal setting potentially overlooks the non-stationary task environments in the real world, where data observations of new views are accumulated over time. Intuitively, once a new view is collected, it is expected to learn it immediately rather than wait for a few new views to occur and learn them together. Second, another question arises of how to upgrade the well-trained multi-view model for each newly collected view. One straightforward solution would be to maintain a training exemplar buffer composed of continual views of data and then merge them again and again. However, it is computationally inefficient to train the model from scratch cumulatively \cite{yin2021incremental, wan2022continual}, and becomes infeasible when a buffer is not allowed due to storage constraints, data privacy concerns, and time consumption \cite{shokri2015privacy}. Besides, it is a common real-world problem that newly arrived views might contain fresh and unseen classes \cite{li2023CLSNet}. Unfortunately, prior work on multi-view classification potentially fails to incrementally learn views of such new classes, revealing its scalability problem.

The above-unsolved issues motivate us to investigate a new paradigm termed multi-view class incremental learning (MVCIL) that can address a broader range of realistic scenarios. In this paradigm, the model learns from multiple views of data while incrementally learning new classes, leading to a more scalable and adaptable learning system \cite{lesort2020continual, delange2021clreview}. For example, we have a rescue robot equipped with multiple sensors, such as cameras, lidar, and thermal imaging, to navigate through a disaster zone and identify victims. The objective is to learn how to recognize different objects, such as debris, furniture, and human bodies, and identify victims in need of rescue. Since there is a delay in data collection for each sensor, the robot processes the camera view first, followed by the lidar and thermal imaging views as they become available. Additionally, the robot must recognize new objects as they are encountered, such as a new type of furniture that arrives from each sensor asynchronously, as formulated later in the preliminary section \ref{Problem_definition}. 

\begin{figure}[tbp]
\centering
\includegraphics[width=1.0\columnwidth]{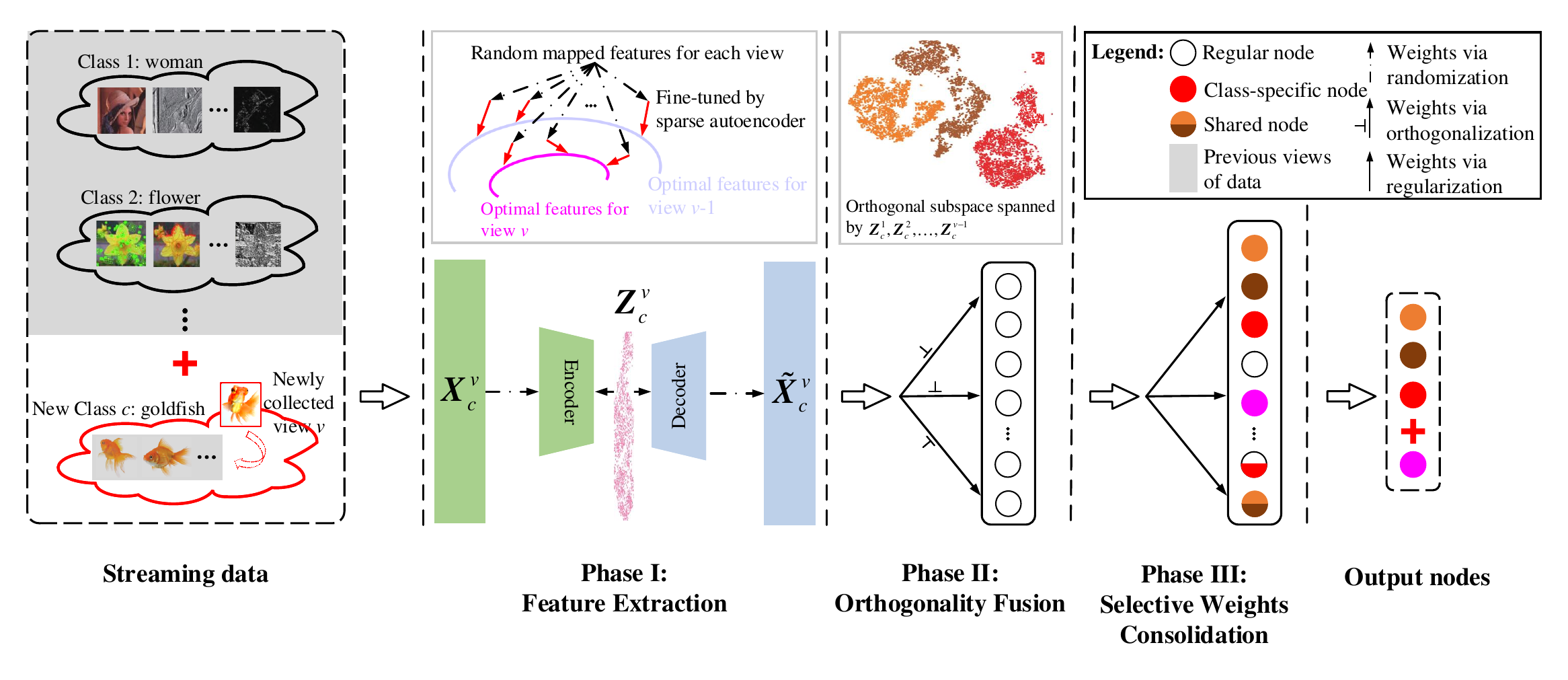}
\caption{The proposed MVCNet pipeline. Given the newly collected view $v$ in the latest class $c$, Phase I employs randomization-based representation learning to circularly yield compact features; Phase II fuses the new view and optimizes corresponding weights in the direction orthogonal to the subspace spanned by extracted features; and Phase III exerts regularization constraints on the updates of important weights and makes predictions.}
\label{Fig_MVCIL}
\end{figure}

To obtain a good grasp of the MVCIL paradigm, this paper adopts the viewpoint on incremental learning \cite{van2022three} and embeds its terminology into the multi-view classification tasks. On this basis, we propose a novel classification network for addressing a continual stream of views, dubbed MVCNet. Instead of only working on stationary independent and identically distributed (IID) multi-view classification data, our algorithm can consecutively learn views from new classes emerging in a sequence of tasks over time, without storing and revisiting previous views of data. An overview of the proposed MVCNet is shown in Fig. \ref{Fig_MVCIL}, which encompasses three phases. Specifically, assuming multiple views belonging to a class are presented sequentially, (1) we first develop randomization-based representation learning, a sparse autoencoder-based feature extraction technique with random weights, to guarantee their separate view-optimal working states. Once a newly collected view is available, it extracts features without depending on task-specific supervision, such that its meaningful representations can be used as input to a stand-alone supervised learning algorithm; (2) Then, we integrate the new view in the orthogonality fusion subspace spanned by extracted features, which leverages the view correlations across previous and current ones while still allowing for some degree of freedom to exploit future views; (3) Finally, we measure parameter importance and selectively regularize weights on the decision layer to alleviate catastrophic forgetting. This strategy only allows the new class to change its connecting weights that deem unimportant for old classification tasks. It is worth mentioning that MVCNet can automatically discriminate between all classes seen so far without knowing task identities at inference time.

The investigation presented can be regarded as an extended study of the existing MVL families, providing an innovative idea and direction for future research on MVL. In summary, the main contributions of this paper include:

\begin{itemize}
    \item MVCNet considers a common yet challenging paradigm, which is applicable to process continual views emerging over time in an open-ended environment and not limited to a prepared dataset with fixed classes. To the best of our knowledge, this is not well-studied in the MVL field.
    \item Without storing data of past views, we present orthogonality fusion to integrate the newly collected view, followed by selective weight consolidation to accommodate a new class. This enables forward transfer from the history information to new concepts without an entire retraining cycle, saving a large number of computational resources.
    \item We conduct extensive experiments to demonstrate the effectiveness of MVCNet on synthetic and real-world datasets. The proposed approach achieves competitive results in the incremental learning scenario while existing multi-view classification algorithms suffer from serious performance degradation.
\end{itemize} 

The remainder of this paper is organized as follows. The related work about multi-view learning and incremental learning is briefly reviewed in Section \ref{Sec_Pre}, followed by the problem definition. In Section \ref{Sec_Method},  we present the methodology of our multi-view class incremental learning algorithm. Section \ref{Sec_Exp} presents comparative experiments to evaluate the superiority of the proposed method. Finally, we conclude our paper and suggest some directions for future research in Section \ref{Sec_Con}.  

\newpage

\section{Preliminaries}
\label{Sec_Pre}

\subsection{Related work}
\textbf{Multi-View Learning.} 
Learning data with different views, such as multiple sensors, modalities, or sources, has proven effective in a variety of tasks \cite{zhang2019feature, houthuys2021tensor, liu2021one, liu2022deep, liu2019absent, zhang2021multilevel}. \textcolor{Blue}{A substantial body of work has extensively studied MVL methods for exploring data correlation across multiple views, from which we discuss some recent representative approaches for three common MVL tasks. (1) \textit{Multi-view clustering} partitions similar objects into the same class by using fused multi-view information. For instance, CoSFL \cite{yan2023collaborative} integrates structure learning and feature learning within a unified framework. SRL \cite{huang2023smooth} utilizes a low-pass graph filter to acquire smooth representations, which are subsequently embedded in the multi-view clustering to facilitate downstream clustering tasks. By employing a one-step strategy, UOMvSC \cite{tang2022unified} combines the spectral embedding with $k$-means to obtain a unified graph, while E$^2$OMVC \cite{wang2023efficient} simultaneously generates a common latent representation and a clustering indicator matrix for multi-view data. (2) \textit{Multi-view representation learning} aims to extract comprehensive feature representations that can be readily utilized by off-the-shelf algorithms. In this context, CMRL \cite{zheng2023comprehensive} introduces a degeneration mapping model and low-rank tensor constraint to fully exploit useful information contained in both the feature representations and subspace representations from multiple views, thereby effectively harnessing consistency and complementarity across views. (3) \textit{Multi-view classification} recognizes instances based on multiple views or representations of the same data by leveraging the label information. A classification model called RED-Nets \cite{fu2021red} first integrates different views and then redistributes them into multiple pseudo views, breaking the barriers among original views. GPVLM \cite{li2021asymmetric} is a generative and non-parametric model that incorporates label information into the generation process through Gaussian Process transformation.} Besides, there has been an increasing focus on contrastive learning methods \cite{tian2020contrastive, liu2023simple}. More recently, we have seen a shift towards real-world scenarios, including how to address incomplete views \cite{liu2018late, zhang2021one}, trusted views \cite{han2022trusted, liu2022trusted}, and streaming views \cite{zhou2019incremental, yin2021incremental}.

\textbf{Incremental Learning.}
The ability to learn distinct tasks consecutively is referred to as incremental learning, a synonym for continual learning and lifelong learning \cite{lesort2020continual, delange2021clreview, le2022uifgan}. There are three incremental learning scenarios, depending on how the aspect of the data that changes over time relates to the function or mapping \cite{van2022three, van2020replay}. (1) \textit{Task-incremental learning} (TIL) allows a model to produce task-specific components, which are structured by additional information such as the provision of task identities at inference time \cite{kang2022forget}; (2) \textit{Domain-incremental learning} (DIL) shares the identical set of outputs for all tasks and needs to accommodate the ever-changing input distributions \cite{van2020replay, li2022domain}. (3) \textit{Class-incremental learning} (CIL) tackles the common real-world problem of incrementally learning new classes of objects \cite{ICCV2019IL2M, li2023IF2Net}. Among them, the last one is mostly related to our work. Contrary to typical machine learning methods that have access to all training data at the same time, CIL places a single model in a dynamic environment where the data arrives in a sequence and the underlying distribution of the data changes over time. Consequently, non-IL methods suffer from a sharp performance degradation on previously learned tasks, a phenomenon known as \textit{catastrophic forgetting} \cite{mccloskey1989catastrophic}. Existing studies to address this problem can be broadly segmented into three categories: replay-based methods \cite{van2020replay, shin2017replay}, regularization-based methods \cite{PNAS2017EWC, li2023MRNet}, and parameter isolation-based methods \cite{serra2018overcoming, gao2022efficient}. Readers may refer to our recent work \cite{li2023CRNet} for further understanding of incremental learning.

\textcolor{Blue}{Among the existing methods that overlap multi-view learning with incremental learning, our work somewhat draws a parallel to Mv-TCNN \cite{li2020continual} and MVCD \cite{yang2022multi}. However, we would still like to emphasize our novelty and superiority compared with these methods. (1) Mv-TCNN first augments training data with different multi-view functions, rather than taking a continual stream of views into consideration; It then trains an expert network for each new task, rendering its model expansion grow linearly with the number of tasks, while our model size maintains relatively unchanged during sequential training. (2) MVCD designs the correlation distillation losses from three specific (i.e., channel-wise, point-wise, and instance-wise) views in the feature space of the object detector, making it less scalable to streaming views by design. By contrast, our method overcomes this constraint by allowing for an unlimited number of views and classes. Traditional incremental learning methods are also related, which involve either dynamically constructing networks itself \cite{dai2019stochastic, dai2021hybrid, dai2019data} or progressively learning a multi-view task \cite{ruvolo2013ella, chen2014topic}. However, these methods fail to address catastrophic forgetting. For example, some prior work simply extends incremental learning to multi-task multi-view learning \cite{li2017lifelong, sun2021continual} or single-task multi-view clustering \cite{zhou2019incremental, yin2021incremental, wan2022continual}, much akin to online learning \cite{hoi2021online} if we treat data of each view as a batch for training. Therefore, our work differs significantly in terms of motivation and methodology.}

\subsection{Problem definition} \label{Problem_definition}
We now define the MVCIL paradigm formally. To facilitate the description, we assume that (1) multiple views are collected by different sensors with varying processing response times; and (2) each class is learned one by one. For example, multiple views belonging to the first class are presented sequentially, followed by that of the second and then third classes. Without loss of generality, the extension to other cases is straightforward. Given the newly collected view $v$ in the latest class $c$ ($c=1,2,\dots$), as shown in Fig. \ref{Fig_MVCIL}, we have access to the supervised learning datasets 
\textcolor{Blue}{$\{(\bm{X}_c^v, \bm{Y}_c)|\bm{X}_c^v\in\mathbb{R}^{d_c^v\times N_c}, \bm{Y}_c\in\mathbb{R}^{1\times N_c}\}$ but none of the past views of data $\{\bm{X}_c^1, \dots, \bm{X}_c^{v-1}\}$, where $N_c$ and $d_c^v$ are the corresponding number of tasks and the dimension, respectively.}
This corresponds to a well-trained multi-view classification model $\mathcal{M}(\bm{X}_c^v;\bm{\theta}_c^{1:v})$, parameterized by its connection weights $\bm{\theta}_c^{1:v}$ over existing views $1\sim v$ of this class. Considering a newly arriving view $\bm{X}_c^{v+1}$, the fusion objective is to train an updated model $\mathcal{M}(\bm{X}_c^{v+1};\bm{\theta}_c^{1:v+1})$ by leveraging the view correlations across all views, without revisiting past views of data; More generally, when a new (first) view represented an unseen class $\{\bm{X}_{c+1}^1,\bm{Y}_{c+1}\}$ is available, the prediction objective is to endow our model to accommodate the newly emerging class $c+1$. Similarly, $\mathcal{M}(\bm{X}_{c+1}^1;\bm{\theta}_{c+1}^1)$ needs to remember how to recognize well the classes seen so far by fusing the incoming views $\bm{X}_{c+1}^{v}$ ($v\ge 2$) belong to this class, i.e., the updated model $\mathcal{M}(\bm{X}_{c+1}^{v};\bm{\theta}_{c+1}^{1:v})$. At inference time, the resulting model would be fed data of views from any of these classes and make decisions without providing any task-specific identity information.

\section{The proposed MVCNet} 
\label{Sec_Method}
This section investigates multi-view classification in the incremental learning setting where continual views emerge in a sequence of classes over time, instead of being ideally collected in advance. To this end, we develop a novel approach called MVCNet for multi-view class incremental learning. The basic idea is to keep network weights relative to each newly arriving view without infringing upon others as there is no access to previous views of data in each training session. In detail, MVCNet precisely controls the parameters of a well-trained model without being overwritten, ensuring that the decision boundary learned fits views of new classes while retaining its capacity to recognize previously learned ones. The following explains how the involved three-step pipeline works (see Fig. \ref{Fig_MVCIL}), including view-level feature extraction, inter-view orthogonality fusion, and inter-class prediction.

\subsection{Randomization-based representation learning}
An autoencoder is an unsupervised neural network that learns latent representations through input reconstruction \cite{vincent2010stacked}. \textcolor{Blue}{During the coding process, meaningful information can be retained in the learned representations or network parameters, which are usually optimized by the stochastic gradient descent (SGD) method. Instead, we propose to utilize a parameter-free sparse optimization as an alternative solution for obtaining parameters of the representation learning. In this way, the representation learning retains the merit of fast convergence and potentially fits for processing sequential data, e.g., without extra care to the learning rate and batch size in SGD for addressing streaming views. However, either using sparse optimization or SGD, the resultant autoencoder is prone to catastrophic forgetting when presented with a continuous stream of views. This occurs because a previously optimized parameter set of old views/tasks would be overridden during the learning process of new views. Consequently, the parameter-overwritten autoencoder would only process the most recently collected view well. To tackle this issue, we challenge common practice to emphasize the extracted feature representations that could capture the intrinsic structure of streaming views, while deemphasizing the importance of autoencoder parameters through random assignment of input weights (and biases). In other words, the weights in encoder-decoder layers can be transitional or disposable, until pushing random mapping features for each view back to their separate view-optimal working states. We now elaborate on how randomization-based representation learning works for view-level feature extraction.
}

\textcolor{Blue}{
During randomization-based representation learning, the encoder and decoder are both constructed with a \textit{single-hidden linear} layer. Specifically, the encoder $f_{\bm{\theta}}(\cdot)$ is parameterized by the randomly assigned parameter set $\bm\theta = \{\bm w, \bm b\}$ to obtain a random mapping feature $\bm Z_c^v = f_{\bm{\theta}}(\bm X_c^v)\in\mathbb{R}^{\tilde{d}_c^v\times N_c}$ from input $\bm X_c^v \in\mathbb{R}^{d_c^v\times N_c}$ of view $v$, where $f_{\bm{\theta}}(\bm X_c^v) = \bm w \bm X_c^v + \bm b$. The decoder $g_{\bm{\theta}^{'}}(\cdot)$, parameterized by $\bm\theta^{'} = \{\bm w^{'}, \bm b^{'}\}$, aims to map from the feature back into the original data space via the reconstructed input $\tilde{\bm X_c^v} = g_{\bm{\theta}^{'}}(\bm Z_c^v)$, where $g_{\bm{\theta}^{'}}(\bm Z_c^v) = \bm w^{'} \bm Z_c^v + \bm b^{'}$.} 
Then, randomization-based representation learning seeks to find the parameter sets $\bm{\theta}$ and $\bm{\theta}^{'}$ that minimize the reconstruction error on a training set of examples by considering the following sparse optimization problem:

\begin{equation}\label{SAE}
    \begin{split}
        \mathop{\arg\min}\limits_{\bm{\theta}, \bm{\theta}^{'}}&: \Vert g_{\bm{\theta}^{'}}(f_{\bm{\theta}}(\bm X_c^v)) - \bm X_c^v \Vert^2 + \Vert \bm{\theta} \Vert_{\ell_1}   \\
	s.t.&:  \bm{\theta}^{\mathrm{T}} - \bm{\theta}^{'} = \bm{0}
    \end{split}
\end{equation}

\noindent where $\ell_1$-norm penalty term is used to generate more sparse and compact features of the input $\bm X_c^v$. For the purpose of view-level feature extraction, 
Eq. (\ref{SAE}) can be equivalently considered as the following general problem:

\begin{equation}\label{Theta}
    \begin{split}\arg\min_{\bm\theta^{'}}&: p(\bm\theta^{'}) + q(\bm\theta^{'})  \\
    \end{split}
\end{equation}

\noindent where $p(\bm\theta^{'}) = \Vert g_{\bm\theta^{'}}(\bm Z_c^v) - \bm X_c^v \Vert^2$ and $q(\bm{\theta}^{'}) = \Vert\bm{\theta}^{'} \Vert_{\ell_1}$. Eq. (\ref{Theta}) can be solved by extending the fast iterative shrinkage-thresholding algorithm (FISTA) \cite{beck2009fast} to view-level feature extraction, as described in \textbf{Algorithm \ref{alg_1}}.

By computing the iterative steps, we denote $\bm\theta^{*}$ ($\bm\theta^{*} \leftarrow \bm\theta^{'*\mathrm{T}}$) as the solution to Eq. (\ref{SAE}). Once we obtain the view-optimal feature $f_{\bm\theta^{*}}(\bm X_c^v)$, the involved weights $\bm\theta^{*}$ in the encoding layer can be reset as the initially randomized weights (and biases) for subsequent views, as depicted on the top panel of Phase I in Fig. \ref{Fig_MVCIL}. This is comparable to adaptively fine-tuning random mapping features for each view back to their separate view-optimal working state. During the above process, randomization-based representation learning is designed for unsupervised network parameter optimization, which does not require explicit label information. Therefore, the randomly assigned weights (and biases) can be reused to reproduce the optimal representations of each arriving view. Meanwhile, to diversify the counterparts of output representations, we optionally construct several groups of $\bm Z_c^v$ at the same time. In detail, the encoder randomly generates $n$ groups of node blocks, with each group $L$ nodes. Then, one can recursively obtain the fine-tuned features by concatenating all groups of node blocks. This is beneficial to disentangle the hidden structure within every view of data.

\begin{algorithm}[tbp]
\caption{Extending FISTA to view-level feature extraction}
\label{alg_1}
\begin{algorithmic}[1]
    \Require 
    The newly collected view $\bm{X}_c^v$ belonging to a class $c$;
    \Statex \quad\quad randomly assigned parameter set $\bm\theta = \{\bm w, \bm b\}$ of encoder $f_{\bm{\theta}}(\cdot)$;
    \Statex \quad\quad maximum times of iteration $K$.
     
    \State Initialize $\bm y_1 \leftarrow \bm\theta^{'}$, $t_1 \leftarrow 1$, and $k \ge 1$;
    \State Calculate the Lipschitz constant $\xi$ of the gradient of the convex
    \Statex function $\nabla p(\bm\theta^{'})$ in Eq. (\ref{Theta});
    \Repeat 
        \State Update $\bm\theta^{'}_k \leftarrow l_{\xi}(\bm y_k)$, where $l_{\xi}$ is given by: 
        \Statex \quad~ $l_{\xi} = \arg\min_{\bm\theta^{'}_k}\big\{\frac{\xi}{2} \Vert \bm\theta^{'}_k - (\bm\theta^{'}_{k-1} - \frac{1}{\xi}\nabla p(\bm\theta^{'}_{k-1})) \Vert^2 + q(\bm\theta^{'}_k)\big\}$;
        \State Set the coefficient $t_{k+1} \leftarrow \frac{1+\sqrt{1+4t_{k}^2}}{2}$;
        \State Calculate $\bm y_{k+1} \leftarrow \bm\theta^{'}_k + \frac{t_{k}-1}{t_{k+1}}(\bm\theta^{'}_k - \bm\theta^{'}_{k-1})$;
    \Until{$k$ satisfies termination criteria $K$}  
    \State Set $\bm\theta^{'*} \leftarrow \bm y_{k+1}$ and $\bm\theta^{*} \leftarrow \bm\theta^{'*\mathrm{T}}$;
    \State Obtain the view-optimal feature $f_{\bm\theta^{*}}(\bm X_c^v)$;
    \State Reset $\bm\theta \leftarrow \bm\theta^{*}$ by the initial randomly assigned $\{\bm w, \bm b\}$;
    \State \textbf{Return} View-optimal feature $f_{\bm\theta^{*}}(\bm X_c^v)$ ($c=1,2,\dots$).
\end{algorithmic}
\end{algorithm}

\subsection{Orthogonality fusion}
\label{Orthogonality_fusion}
\textcolor{Blue}{The motivation for introducing orthogonal fusion is twofold. First, the general idea of view fusion is to exploit the complementary and consensus information assuming that all data views are static and can be simultaneously accessed. However, this largely overlooks the non-stationary task environments in real-world scenarios where new view data observations accumulate over time. Second, cumulatively storing each newly collected view and retraining them from scratch may be memory-inefficient and time-consuming; and importantly, it raises practical concerns such as privacy and security issues \cite{shokri2015privacy}, which are common in domains like federated learning. With these considerations, inspired by the recent advances of orthogonal gradient descent that keep the input-to-output mappings untouched \cite{NMI2019OWM, li2021gopgan}, we present an orthogonality fusion strategy for a continual stream of views. This offers a promising solution to fusing streaming views without the necessity of storing data or features from previously learned views when addressing a new one, thereby contributing to an innovative concept for the MVL community.} 

Specifically, based on the view-optimal feature $\bm Z_c^v\in\mathbb{R}^{\tilde{d}_c^v\times N_c}$ from the randomization-based representation learning, we denote $\bm W_c^{1:v}\in\mathbb{R}^{\sum_v\tilde{d}_c^v\times l}$ as the connecting weights with $l$ nodes sequentially optimized over earlier views $1 \sim v-1$ and $\bm Z_c=\{\bm Z_c^1, \dots, \bm Z_c^{v-1}\}\in\mathbb{R}^{\sum_v\tilde{d}_c^v\times N_c}$ as a collection of extracted features so far. \textcolor{Blue}{Unlike the conventional gradient decent method, the weights $\bm W_c^{1:v}$ are updated only in the direction orthogonal to the subspace spanned by $\bm Z_c$ such that the adaptation keeps previously learned mappings intact. In this way, the orthogonality fusion process not only leverages complementary and consensus information among them but also allows for some degree of freedom to exploit future views in a cost-effective manner.} To this end, we introduce the orthogonal projection matrix $\bm{P}_c^{1:v}\in\mathbb{R}^{\sum_v\tilde{d}_c^v\times \sum_v\tilde{d}_c^v}$ for modulating gradient of $\bm W_c^{1:v}$, a nonzero projector satisfying:

\begin{align}
	\bm{P}_c^{1:v} &= \bm{I} -\bm{Z}_c(\bm{Z}_c^\mathrm{T}\bm{Z}_c+\alpha\bm{I})^{-1}\bm{Z}_c^\mathrm{T} \label{P_form} \\
	\bm W_c^{1:v} &= \bm W_c^{1:v-1}-\eta \bm{P}_c^{1:v}\Delta \bm W_c^{1:v} \label{OF}
\end{align}



\noindent where $\eta$ is the learning rate, $\Delta \bm W_c^{1:v}\in\mathbb{R}^{\sum_v\tilde{d}_c^v\times l}$ is the corresponding gradient, and $\alpha \ge 0$ is the orthogonality coefficient. Among them, the hyper-parameter $\alpha$ plays a pivotal role in implementing the orthogonality fusion technique. 

When $\alpha = 0$, each row of $\bm{P}_c^{1:v}$ is orthogonal to the space spanned by the collection of extracted features $\bm Z_c$ which we refer to as \textit{orthogonal subspace}. This implies that for every view feature $\bm Z_c^j$ in $\bm Z_c$, we have $\bm{P}_c^{1:v} \bot \bm Z_c^j$ $(j=1,2,\dots,v-1)$. Then, as the number of fused views increases, the collection matrix $\bm Z_c$ tends to become full rank such that there exists no nonzero projector $\bm{P}_c^{1:v}$ given by Eq. (\ref{P_form}). When $\alpha$ gets bigger, the projector $\bm{P}_c^{1:v}$ will pay more attention to the dominant components of $\bm{P}_c^{1:v} \bm Z_c^j$ to be zero while disregarding the non-dominant components, which is significant for the fusion freedom (margin) of orthogonality in exploiting subsequent views. This aligns with the discussion in the literature \cite{li2021gopgan} where the orthogonality is used for generative tasks. Actually, $\alpha$ is a trade-off between the ‘stability’ (leading to intransigence) and ‘plasticity’ (resulting in forgetting). When $\alpha$ is large, $\bm{P}_c^{1:v}$ is biased towards identity matrix $\bm I$, which focuses more on fitting new views. As we reduce $\alpha$ towards 0, it firmly maintains the information of past views but faces the loss of plasticity to learn new views, especially when we have a lot of views such that $\bm Z_c^j$ is close to full rank. In a nutshell, determining $\alpha$ properly, e.g., reducing its value by a relatively small constant, will improve the performance of orthogonality fusion (see Section \ref{Parameter_analysis}).

Since it is too strict to incrementally guarantee the existence of orthogonal subspace maintained by $\bm{P}_c^{1:v} \bot \bm Z_c^j$ $(j=1,2,\dots,v-1)$, we approximate the subspace by employing an iterative method, which also breaks the limitation of storing features from all past views. We embed this strategy into the recursive least squares (RLS) \cite{haykin2002adaptive} to recursively compute $\bm{P}_c^{1:v}$ with only $\bm Z_c^v$ instead of the collection $\bm Z_c$, which is flexible and easy to implement. Specifically, we treat the first $k$ feature vectors of $\bm Z_c^v$ as a mini-batch, e.g., $\bm Z_c^v(k)=[\bm z_{c,1}^v,\bm z_{c,2}^v,\dots,\bm z_{c,k}^v]\in\mathbb{R}^{\tilde{d}_c^v\times k}$. In this way, $\bm Z_c^v(k+1)\bm Z_c^v(k+1)^\mathrm{T}=\bm Z_c^v(k)\bm Z_c^v(k)^\mathrm{T}+ \bm z_{c,k+1}^v\bm z_{c,k+1}^{v\mathrm{T}}$. Similarly, We denote $\bm{P}_c^{1:v}(k)\in\mathbb{R}^{\tilde{d}_c^v\times \tilde{d}_c^v}$ as the iterative form that corresponds to $\bm Z_c^v(k)$. Hence, Eq. (\ref{P_form}) can be converted into the following recursive formula:

\begin{equation}\label{P_k}
\bm{P}_c^{1:v}(k+1)=\alpha(\bm Z_c^v(k)\bm Z_c^v(k)^\mathrm{T} +  \bm z_{c,k+1}^v\bm z_{c,k+1}^{v\mathrm{T}} + \alpha\bm{I})^{-1}
\end{equation}

\noindent To make the computation of the projector more efficient, we further simplify Eq. (\ref{P_k}) using the Woodbury matrix identity \cite{golub2013matrix}, as is presented below. 






\begin{equation}\label{P_kk}
\begin{split}
&\bm{P}_c^{1:v}(k+1)=\bm{P}_c^{1:v}(k) -\bm{Q}_c^{1:v}(k)\bm z_{c,k+1}^{v\mathrm{T}} \bm{P}_c^{1:v}(k)\\
&\bm{Q}_c^{1:v}(k)=\bm{P}_c^{1:v}(k)\bm z_{c,k+1}^v/\big(\alpha\bm{I}+\bm z_{c,k+1}^{v\mathrm{T}}\bm{P}_c^{1:v}(k)\bm z_{c,k+1}^v\big)
\end{split}
\end{equation}

\noindent Eq. (\ref{P_kk}) replaces the computation of the matrix-inverse operation with the iterative implementation. Meanwhile, it only requires the currently learned $\bm z_{c,k+1}^v$, a batch from $\bm Z_c^v$, to perform the orthogonality fusion on each newly arriving view. This strategy preserves the correlative and complementary information and then provides sufficiently discriminative features for subsequent classification.

\subsection{Selective weight consolidation}
In real applications, new classes often emerge after the classification model has been well-trained on a dataset with fixed classes. Fine-tuning the model with only data from new classes will lead to the well-known catastrophic forgetting phenomenon. On the other hand, it is rather ineffective to train the model from scratch with all data from both old and new classes. 

As a method of choice for the problem-solving, we first briefly introduce Elastic Weight Consolidation (EWC) \cite{PNAS2017EWC}, which employs an approximate Bayesian estimation to regularize weight updates, i.e.,

\begin{equation}\label{EWC}
    \mathcal{L(\theta)} = \mathcal{L}_c(\theta) + \frac{\mu}{2}\sum_{t=1}^{c-1}\sum_{l}\sum_{i,j}\mathcal{F}_{i,j}^{t, l}(\theta_{i,j}^l-\theta_{i,j}^{t, l})^2
\end{equation}

\noindent where $\mathcal{L}_c(\theta)$ is the loss for class/task $c$, $\mu$ is the regularization coefficient, $\sum_{t=1}^{c-1}$ refers to the past $c-1$ classes sequentially presented, $\sum_{l}$ exerts the layer-wise regularization, $\sum_{i,j}$ covers each connection weight between two adjacent layers, and $\bm {\mathcal{F}}^t$ is a Fisher information matrix indicating its importance at each layer $l$ with respect to the previous class $t$, i.e.,

\begin{equation}\label{EFIM}
\bm {\mathcal{F}}^t=\frac{1}{N_t}\sum_{i=1}^{N_t}\nabla_{\bm{\theta}}\log p_{\bm{\theta}}(\bm{x}_i)\nabla_{\bm{\theta}}\log p_{\bm{\theta}}(\bm{x}_i)^\mathrm{T}
\end{equation}

\noindent Unlike $L_2$ regularization that acts equally on all parameters, it treats parameters in each layer as having a different level of importance. Therefore, EWC keeps a quadratic penalty for each previous task such that it could constrain the parameters not to deviate too much from those that were optimized. 

However, the original EWC algorithm forces a model to remember older tasks more vividly by double counting the data from previous tasks, i.e., for each layer, the accumulation of Fisher regularization would over-constrain the network parameters and impede the learning of new tasks; for the entire EWC network, it suffers from a build-up of such inflexibility that is proportional to the number of layers. On the other hand, the number of regularization terms grows linearly with the number of tasks, which renders it non-scalable for a large number of tasks. To tackle these issues, the following elaborates on two improvements to EWC that we refer to as \textit{selective weight consolidation}. Specifically, (1) we replace $\theta_{i,j}^t$ with $\theta_{i,j}^{c-1}$ by anchoring at the weights associated with the latest task since the most recently learned weights have inherited the previous ones. Correspondingly, the re-centering is achieved along with a running sum of the Fisher regularization, yielding only a single term; and (2) we build on top of the orthogonality fusion and only apply the consolidation to the final output layer for decision-making, i.e., our approach requires only the maintenance of a single Fisher regularization across a sequence of tasks. In this way, the systematic bias towards previously learned tasks can be effectively counterbalanced by controlling the regularization coefficient $\mu$. Therefore, in our MVCNet, the loss of selective weight consolidation serving for the final classifier is:

\begin{equation}\label{SWC}
    \mathcal{L(\theta)} = \mathcal{L}_c(\theta) + \frac{\mu}{2}\sum_{i,j}(\sum_{t=1}^{c-1}\mathcal{F}_{i,j}^t)(\theta_{i,j}-\theta_{i,j}^{c-1})^2
\end{equation}

\textcolor{Blue}{
To gain insights into our selective weight consolidation, we want to emphasize its advantages in terms of \textit{parameter efficiency}. (1) The original EWC, as formulated in Eq. (\ref{EWC}) needs to gradually accumulate quadratic penalty terms, resulting in a substantial linear increase in computational demands with the number of tasks. By contrast, our improved strategy can be updated by a moving sum $\sum_{t=1}^{c-1}\mathcal{F}_{i,j}^t$, e.g., one can maintain the most recently learned weights $\theta_{i,j}$ but throw away all previous ones $\theta_{i,j}^{t} (t=1,2,\dots, c-1)$. Therefore, Eq. (\ref{SWC}) used in our method keeps the number of parameters relatively unchanged during sequential training. (2) Empirically, we observe that EWC is highly susceptible to the regularization coefficient $\mu$ for different settings (e.g., learning rate) and benchmark datasets as revealed by \cite{li2023CRNet}, which has been well tackled in our MVCNet. In particular, the re-centering built upon a single regularization term facilitates the counterbalance of systematic bias towards previously learned tasks, as demonstrated in the experiments. 
}

Owe to selective weight consolidation, as shown in Fig. \ref{Fig_MVCIL}, some nodes in the penultimate layer are reserved to be speciﬁc for that task while other nodes can be shared and reused among multiple tasks based on their importance towards the task. This allows incoming tasks to use the previously learned knowledge while maintaining plasticity. It is worth mentioning that the above decision-making process requires no access to previous views of data. In addition, it can automatically discriminate between all classes seen so far without knowing task identities at inference time. We summarize the algorithmic implementation procedure in \textbf{Algorithm \ref{alg_2}}.

\begin{algorithm}[tbp]
\caption{Training procedure of the proposed MVCNet}
\label{alg_2}
\begin{algorithmic}[1]
    \Require 
    The newly collected data $\{\bm{X}_c^v, \bm{Y}_c\}$ of view $v$ belonging to the latest class $c$; 
    \Ensure 
    Network predictions on classes seen so far.
    \For{$c=1, 2,\dots,C$}
        \For{$v=1, 2,\dots,V$}
        \State Extract compact representations $\bm Z_c^v$ with Eq. (\ref{SAE});
        \State Compute orthogonal projection matrix $\bm{P}_c^{1:v}$ with Eq. (\ref{P_kk});
        \State Perform orthogonality fusion with Eq. (\ref{OF});
        \State Selectively consolidate weights of read-out layer with Eq. (\ref{SWC});
        \EndFor
        \If{$c<C$}
            \State Obtain weight importance matrix $\bm {\mathcal{F}}^c$ with Eq. (\ref{EFIM});
        \EndIf
    \EndFor
\end{algorithmic}
\end{algorithm}

\subsection{\textcolor{Blue}{Time complexity analysis}}
\textcolor{Blue}{
In this section, we analyze the time complexity of the proposed MVCNet by phase. The time complexity of Phase I lies in solving $\bm\theta^{*}$. After ignoring the simple matrix multiplication and addition of intermediate variables, it costs $\mathcal{O}(K)$ with $K$ being the times of iteration for sparse optimization; In Phase II, the orthogonal projector $\bm{P}_c^{1:v}(k)$ is the main computational cost and its complexity is $\mathcal{O}(\tilde{d}_c^{v2} + \tilde{d}_c^{v3})$ per iteration, where $\tilde{d}_c^v$ is the dimension of extracted features; The time complexity in Phase III for updating $\bm {\mathcal{F}}^t$ is $\mathcal{O}(N_cl'^2)$, where $l'$ and $N_c$ denote the number of nodes in the penultimate and decision layers respectively. Since time complexity primarily focuses on understanding how computational time scales with input data size, we additionally report the actual running time (sec.) in Table \ref{Table_5} for a direct comparison with competitors.   
}

\section{Experiments}
\label{Sec_Exp}

\subsection{Experiment setup}
\textbf{Datasets.} Five popular classification datasets of visual tasks are used for comparison. (1) \textbf{COIL-20} \cite{nene1996columbia} contains 1440 grayscale images of 20 objects (72 images per object) under various poses, which are rotated through $360^{\circ}$ and taken at the interval of $5^{\circ}$. (2) \textbf{Animals with Attributes (AwA)} \cite{lampert2013attribute} includes 30475 images of 50 animal subjects. The views (visual representations) used are generated by various deep convolutional neural network-based methods. (3) \textbf{PIE} \cite{gross2007cmu} contains more than 750,000 images of 337 people under various views, illuminations, and expressions. (4) \textbf{MNIST} \cite{lecun1998gradient} is a large collection of handwritten digits. It has a training set of 60,000 examples and a test set of 10,000 examples, with each represented by 28$\times$28 gray-scaled pixels. (5) \textbf{FashionMNIST} \cite{FashionMNIST} shares the same image size, data format and the structure of training and testing splits with the original MNIST.

\textbf{Protocols.} For convenience, we used the nomenclature “Dataset-$C(V)$” to denote a task sequence with $V$ views belonging to each one of the $C$ classes, i.e., the suffix “$C(V)$” indicates that a model needs to progressively fusion $V$ views in each class and incrementally distinguish between a total of $C$ classes seen so far. Hence, there are no overlapping views in and between different classes. The following introduces the resulting synthetic and real-world datasets and their important statistics are summarized in Table \ref{Table_1}.

\begin{table}[htbp]
\caption{Details of the datasets for multi-view class incremental learning used in the experiments.}
    \label{Table_1}
    \centering
    \begin{tabular}{lccc}
    \toprule
    Datasets   & Views & Classes  & Samples \\ \midrule
    COIL-20(4) & 4  & 20   & 1440 $\times$ 1    \\
    COIL-20(8) & 8 & 20  & 1440 $\times$ 1    \\
    AwA-50(2) &2  &50 &10158 $\times$ 2        \\
    PIE-68(3) &3  &68  &680 $\times$ 3        \\
    PMNIST-10(3) & 3 & 10  & 70000 $\times$ 3   \\
    PFMNIST-10(3) & 3 & 10  & 70000 $\times$ 3   \\
    SMNIST-10(3) & 3 & 10  & 70000 $\times$ 3   \\
    SFMNIST-10(3) & 3 & 10  & 70000 $\times$ 3 \\ \bottomrule
    \end{tabular}
\end{table}

\begin{itemize}
    \item COIL-20(4) and COIL-20(8) are drawn by step sizes of $90^{\circ}$ and $45^{\circ}$ from 20 classes in which each class is represented by 4 views and 8 views, respectively.
    \item AwA-50(2) is a subset of the original dataset, which contains 10158 images with two types of deep features extracted with DECAF \cite{krizhevsky2017imagenet} and VGG19 \cite{simonyan2014very} employed in the experiments.
    \item PIE-68(3) is a subset of 680 facial images of 68 subjects employed to evaluate face recognition across poses. Three types of features including intensity, LBP, and Gabor are extracted for the experiment.
    \item PMNIST-10(3) permutates the pixels of the MNIST images in different ways, i.e., three random permutations are generated and applied to these 784 pixels.
    \item PFMNIST-10(3) is a permutated version of FashionMNIST, similar to PMNIST-10(3).
    \item SMNIST-10(3) is synthesized by the randomization-based representation learning with three visual representations per class. 
    \item SFMNIST-10(3) is a synthesized version of FashionMNIST, similar to SMNIST-10(3).
\end{itemize}

\textbf{Compared methods.} We compare our method with the existing MVL classification algorithms: (1) CPM-Nets \cite{zhang2019cpm}; (2) TMC \cite{han2021trusted}; (3) MIB \cite{federici2020learning}; (4) InfoMax \cite{hjelm2018learning}; (5) MV-InfoMax \cite{ji2019invariant}; (6) VAE \cite{alemi2018fixing}; (7) deepCCA \cite{wang2015deep}. Note that we relax the protocol for some methods by satisfying the specific learning requirements as they do not work when directly applying to the MVCIL paradigm. To make the comparison more complete, we also compare with CIL algorithms: (1) EWC \cite{PNAS2017EWC}; (2) PCL \cite{AAAI2021PCL}; (3) OWM \cite{NMI2019OWM}; (4) BiC \cite{wu2019large}; (5) FS-DGPM \cite{deng2021flattening}; (6) GEM \cite{NIPS2017GEM}; (7) LOGD \cite{CVPR2021LOGD}; (8) IL2M \cite{ICCV2019IL2M}. Note that these methods treat each view as a separate incremental learning task.

\textbf{Evaluation Metrics.} We evaluate all considered methods based on the following metrics (higher is better). Average Accuracy (\textbf{Avg Acc}) is the average test classification accuracy on the classes seen so far: $\text{Avg Acc}=\frac{1}{C}\sum_{c=1}^C R_{C,c}$, where $R_{C,c}$ is the test accuracy for class $c$ after training on class $C$; Backward Transfer (\textbf{BWT}) \cite{NIPS2017GEM} is defined as the difference
in accuracy between when a task is first trained and
after training on the final task: $\text{BWT}=\frac{1}{C-1}\sum_{c=1}^{C-1} R_{C,c}-R_{c,c}$. BWT indicates a model's ability in knowledge retention, which complements Avg Acc, e.g., if two models have similar Avg Acc, the preferable one shares the larger BWT. We also report the commonly used offline test classification accuracy (\textbf{Offline Acc}) for MVL methods by offline training as references, which is trained over all classes at once.

\textbf{Implementation Details.} In our experiments, on all datasets, the order of the class is set in a random manner since it is unknown prior to training in the incremental learning context. For all baselines considered, we refer to the original codes for implementation and hyper-parameter selection to ensure the best possible performance. For our method, the hyper-parameters used in our experiment are as follows: $n=30$, $L=20$, and $K=50$ for feature extraction; $\alpha = 1$ and $\eta = 0.01$ for orthogonality fusion; $\mu = 1000$ for selective weight consolidation (see Section \ref{Parameter_analysis} in the parameter analysis for more). Meanwhile, we conduct each benchmark under five independent runs and then report these results' means and standard deviations, implemented with PyTorch-1.8.0 on NVIDIA RTX 3080-Ti GPUs.

\subsection{Results and discussion}
\subsubsection{COIL-20(4) and COIL-20(8)}

\begin{table}[tbp]
\renewcommand{\arraystretch}{0.90}  
\caption{Experimental results on the COIL-20(4) dataset, among which the best results are marked in \textbf{Bold}. $^\dag$ denotes that at least two classes are required for the baseline in each incremental training session.}
\label{Table_2}
\centering
\begin{tabular}{llcc}
\toprule
Description & Method & Avg Acc & Offline Acc  \\ \midrule
\multirow{7}{*}{MVL} 
    & deepCCA$^\dag$  &10.00$\pm$0.00    & 68.89$\pm$0.96    \\
    & InfoMax$^\dag$   &7.45$\pm$0.56    &70.67$\pm$1.25     \\
    & MIB$^\dag$      &10.00$\pm$0.00        &75.37$\pm$3.85  \\
    & MV-InfoMax$^\dag$   &5.97$\pm$0.14 &76.39$\pm$2.83  \\
    & VAE$^\dag$       &9.81$\pm$0.16    &85.21$\pm$2.73      \\
    & CPM-Net     &5.00$\pm$0.00     &86.66$\pm$2.35   \\
    & TMC        &5.00$\pm$0.00 & \bf 95.67$\pm$0.91   \\ \midrule
    & Method & Avg Acc & BWT \\ \cmidrule{2-4}
\multirow{8}{*}{CIL} 
     & EWC   &22.50$\pm$8.13 &-0.36$\pm$0.02   \\
     & PCL   &57.40$\pm$4.14 &-0.09$\pm$0.02   \\
     & FS-DGPM  &71.94$\pm$3.49 &-0.11$\pm$0.02   \\
     & BiC     &73.25$\pm$2.21 &-0.06$\pm$0.01   \\
     & OWM    &75.33$\pm$5.11 &\bf -0.04$\pm$0.01   \\
     & GEM    &77.50$\pm$3.41 &-0.07$\pm$0.03   \\
     & LOGD   &78.15$\pm$4.10 &-0.08$\pm$0.05   \\
     & IL2M   &82.37$\pm$2.97 &-0.12$\pm$0.02    \\ \midrule
MVCIL   & Ours  &\bf 85.56$\pm$3.35 &-0.06$\pm$0.03  \\ \bottomrule
    \end{tabular}
\end{table}

\begin{table}[tbp]
\caption{Experimental results on the COIL-20(8) dataset, among which the best results are marked in \textbf{Bold}. $^\dag$ denotes that at least two classes are required for the baseline in each incremental training session.}
\label{Table_3}
\centering
\begin{tabular}{llcc}
\toprule
Description & Method & Avg Acc & Offline Acc   \\ \midrule
\multirow{7}{*}{MVL} 
    & deepCCA$^\dag$ &10.00$\pm$0.00 &96.67$\pm$3.82   \\
    & VAE$^\dag$  &9.67$\pm$0.27   &92.50$\pm$5.30      \\
    & InfoMax$^\dag$   &10.00$\pm$0.00     &93.54$\pm$2.27     \\
    & MIB$^\dag$   &9.63$\pm$0.32       &94.50$\pm$2.73 \\
    & MV-InfoMax$^\dag$  &9.81$\pm$0.32  &96.10$\pm$3.79   \\
    & CPM-Net     &5.00$\pm$0.00    &96.54$\pm$1.25    \\
    & TMC       &5.00$\pm$0.00     &\bf 97.23$\pm$0.98   \\ \midrule
    & Method & Avg Acc & BWT \\ \cmidrule{2-4}
\multirow{8}{*}{CIL} 
     & EWC   &27.08$\pm$5.70 &-0.26$\pm$0.11   \\
     & PCL   &58.54$\pm$1.47 &-0.08$\pm$0.03   \\
     & FS-DGPM  &69.80$\pm$4.99 &-0.13$\pm$0.06   \\
     & OWM   &78.18$\pm$3.21 &-0.05$\pm$0.01   \\
     & LOGD   &80.15$\pm$3.06 &-0.07$\pm$0.06   \\
     & GEM   &82.81$\pm$1.36 &-0.05$\pm$0.01   \\
     & IL2M   &84.69$\pm$2.36 &-0.08$\pm$0.02    \\ \midrule
MVCIL   & Ours   &\bf 86.76$\pm$2.13 &\bf -0.05$\pm$0.01  \\ \bottomrule
    \end{tabular}
\end{table}

We extensively compare the proposed MVCNet with existing state-of-the-art methods, covering seven MVL baselines and eight CIL baselines. Table \ref{Table_2} reports the results on the COIL-20(4) dataset adapted for MVCIL, where a single model is sequentially fed with a continual stream of 80 views belonging to 20 classes. Our method shows superiority in two measurements. In terms of Avg Acc, we outperform the second-best IL2M by an absolute margin of 3.19\%. In terms of BWT, we achieve a competitive performance on knowledge retention that is slightly inferior to OWM, but our Avg Acc significantly surpasses it. It should be pointed out that most of the MVL methods direct fail to MVCIL paradigm, where they exhibit serious performance degradation compared with their Offline Acc. Actually, without accessing data of past views, the previous knowledge has been totally forgotten, classifying all test images into the newest class. The above observations can also be reflected in the comparison reported in Table \ref{Table_3}, where we extend the length of continual views from 80 to 160.

\subsubsection{AwA-50(2)}
\begin{figure}[htbp]
\centering
\includegraphics[width=1.0\columnwidth]{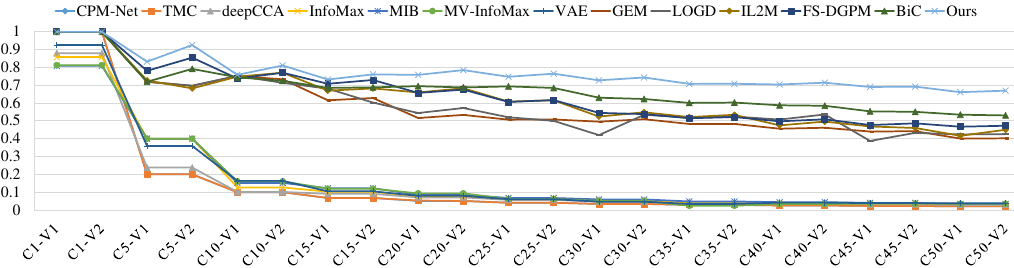}
\caption{Experimental results on the AwA-50(2) dataset.}
\label{Fig_AwA}
\end{figure}

Fig. \ref{Fig_AwA} compares our method with different baselines on the AwA-50(2) dataset, in which we show the test results on classes seen so far after learning views with the step size of every five classes. For example, C10-V2 refers to the average test accuracy in the first 10 classes after sequentially fed with the 2nd view. It can be observed that the proposed method can incrementally fuse complementary and consensus information among continual views well and the final performance tends to be stable. By contrast, the CIL baselines struggle to complete the fusion process such that the overall downward trend is obvious. Meanwhile, MVL methods suffer from serious catastrophic forgetting due to the lack of capability to protect the knowledge learned from previous views, as old knowledge is completely forgotten when learning new concepts. Specifically, the output of the parameter-overwritten model is always the label that has just been learned regardless of the inputs. Although the model has high accuracy in classifying the classes it learned in the end, it has lost its ability to classify the previously learned classes. Therefore, as learning continues, the average test accuracy over all categories it has learned continues to decline. As shown in Fig. \ref{Fig_AwA}, this problem widely exists in MVL models. Interestingly, the initial oscillation may result from the occurrence of the first new class with a single view. As more new classes with different views arrive, the fluctuation gradually tends to flatten out.

\subsubsection{PIE-68(3)}
\begin{figure}[htbp]
\centering
\includegraphics[width=0.7\columnwidth]{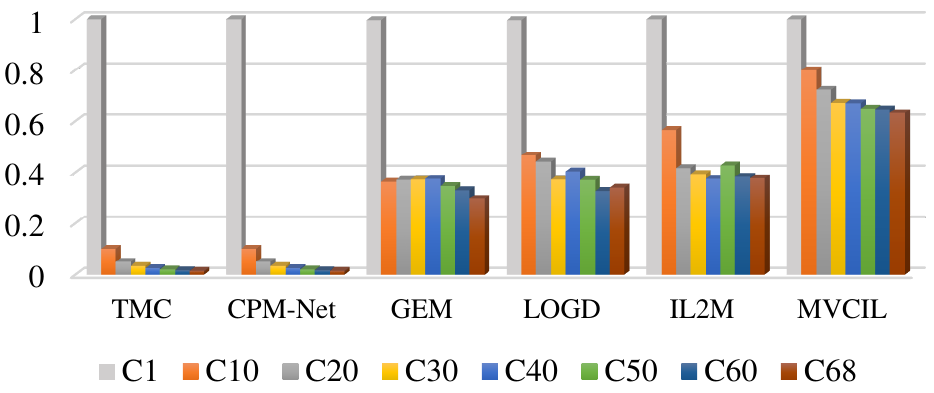}
\caption{Experimental results on the PIE-68(3) dataset.}
\label{Fig_PIE}
\end{figure}

We consider a few-shot case where PIE-68(3) contains only 10 samples for each view. After being sequentially trained with 68 classes, the performance of each method is shown in Fig. \ref{Fig_PIE}, where we only print the results of classes 1, 10, 20, 30, 40, 50, 60, and 68 for brevity. It can be observed that the overall trend is decreasing. Specifically, all the compared methods perform similarly in the first class with three views as this is a conventional single-task learning. Then, there is a sharp decline, especially in the MVL (TMC and CPM-Net) methods, followed by the CIL (GEM, LOGD, and IL2M) methods. By contrast, our algorithm achieves a significant lead on this dataset, indicating the proposed method is also applicable to the data scarcity scenario. 

\subsubsection{PMNIST-10(3) and PFMNIST-10(3)}
PMNIST-10(3) and PFMNIST-10(3) randomly permutate the original 784 pixels in three different orders, thus constituting three different views for each class. The datasets are more diverse among views and thus more difficult to fuse them. On this basis, Table \ref{Table_4} lists the results of sequentially learning 30 continual views from 10 classes. Compared with the second-best method, our proposed method achieves a significant performance gain of 4.59\% and 3.72\% respectively for the PMNIST-10(3) and PFMNIST-10(3) datasets. This implies that our method is able to fuse and learn information well even with substantial intra-view diversity, such as random permutations. Furthermore, for the PFMNIST-10(3) dataset, the Avg Acc of each method generally slipped. This is because the original FashionMNIST is more difficult than MNIST, but our method still shows superiority compared to the selected baselines.

\begin{table}[tbp]
\caption{Experimental results on the PMNIST-10(3) and PFMNIST-10(3) datasets.}
\label{Table_4}
\centering
\begin{tabular}{lcccc}
\toprule
\multirow{2}{*}{Method} &
  \multicolumn{2}{c}{PMNIST-10(3)} & \multicolumn{2}{c}{PFMNIST-10(3)} \\ \cmidrule{2-3} \cmidrule{4-5}
        & Avg Acc & BWT & Avg Acc & BWT \\ \midrule
CPM-Net &10.00$\pm$0.00 &- &10.00$\pm$0.00 &-  \\
TMC     &10.00$\pm$0.00 &- &10.00$\pm$0.00 &- \\
deepCCA &13.47$\pm$0.47 &- &13.16$\pm$0.51 &-     \\
BiC     &79.50$\pm$1.62 &-0.09$\pm$0.03 &70.75$\pm$1.27 &-0.18$\pm$0.01      \\
PCL     &83.42$\pm$0.59 &-0.08$\pm$0.02 &67.43$\pm$0.96  &-0.17$\pm$0.02     \\
IL2M    &83.54$\pm$0.13 &-0.08$\pm$0.01 &75.74$\pm$2.79  &-0.16$\pm$0.01   \\
GEM     &85.15$\pm$3.55 &-0.05$\pm$0.02 &78.03$\pm$3.24  &-0.08$\pm$0.05   \\
FS-DGPM &85.69$\pm$1.14 &-0.09$\pm$0.01 &69.57$\pm$2.19  &-0.20$\pm$0.04     \\
LOGD    &86.85$\pm$2.21 &\bf -0.02$\pm$0.02 &78.66$\pm$0.77  &\bf -0.05$\pm$0.04   \\ \midrule
Ours   &\bf 91.44$\pm$0.43  &-0.07$\pm$0.00 &\bf 82.38$\pm$0.58  &-0.13$\pm$0.01  \\ \bottomrule
    \end{tabular}
\end{table}

\subsubsection{SMNIST-10(3) and SFMNIST-10(3)}
In this experiment, views refer to visual representations generated by our randomization-based representation learning, among which each class is captured by three learned representations. It can be seen from Table \ref{Table_5} that using this view construction method in combination with our framework can achieve over 97\% and 95\% Avg Acc on the SMNIST-10(3) and SFMNIST-10(3) datasets respectively. This indicates that the view constructed using randomization representation-based learning can be compatible with the fusion and decision-making process. In addition, we observe that the other CIL methods can also get better results under this view construction, which to some extent validates the effectiveness of this view construction method in leveraging data diversity. \textcolor{Blue}{Furthermore, we compare the running time with these competitors, which also indicates the competitive superiority in developing a fast MVCIL algorithm.}

\begin{table}[tbp]
\caption{\textcolor{Blue}{Experimental results on the SMNIST-10(3) and SFMNIST-10(3) datasets.}}
\label{Table_5}
\centering
\begin{tabular}{lclcl}
\toprule
\multirow{2}{*}{Method} &
  \multicolumn{2}{c}{SMNIST-10(3)} & \multicolumn{2}{c}{SFMNIST-10(3)} \\ \cmidrule{2-3} \cmidrule{4-5}
        & Avg Acc & \textcolor{Blue}{Time (s)} & Avg Acc & \textcolor{Blue}{Time (s)}  \\ \midrule
CPM-Net & 10.00 &4350.87   & 10.00 &4600.83    \\
TMC     & 10.00 &\bf 145.83    & 10.00 &\bf 155.23     \\
deepCCA & 12.63 &365.18    & 13.47 &364.49         \\
BiC     & 92.26 &2370.15   & 82.97 &2758.25         \\
IL2M    & 92.27 &2036.76   & 84.31 &1986.02         \\
GEM     & 93.69 &529.22    & 89.00 &655.00         \\
LOGD    & 95.32 &17335.48  & 90.13 &17458.46       \\ \midrule
Ours    &\bf 97.88 &208.83 &\bf 95.11 &215.63\\ \bottomrule
\end{tabular}
\end{table}

\subsection{Parameter analysis}
\label{Parameter_analysis}
This section conducts experiments to analyze the sensitivity of hyper-parameters. As discussed in Section \ref{Orthogonality_fusion}, the orthogonality coefficient $\alpha$ needs to be set properly in our method. This is closely related to the trade-off between remembering knowledge of past views well and concentrating more on processing new views. To determine the recommended setting, we make a sensitivity analysis on different levels of $\alpha$ value from $\{10^{-1}, 10^{0}, 10^{1}, 10^{2}\}$.

\begin{figure}[tbp]
    \centering
    \subfloat[]{\includegraphics[height=1.25in]{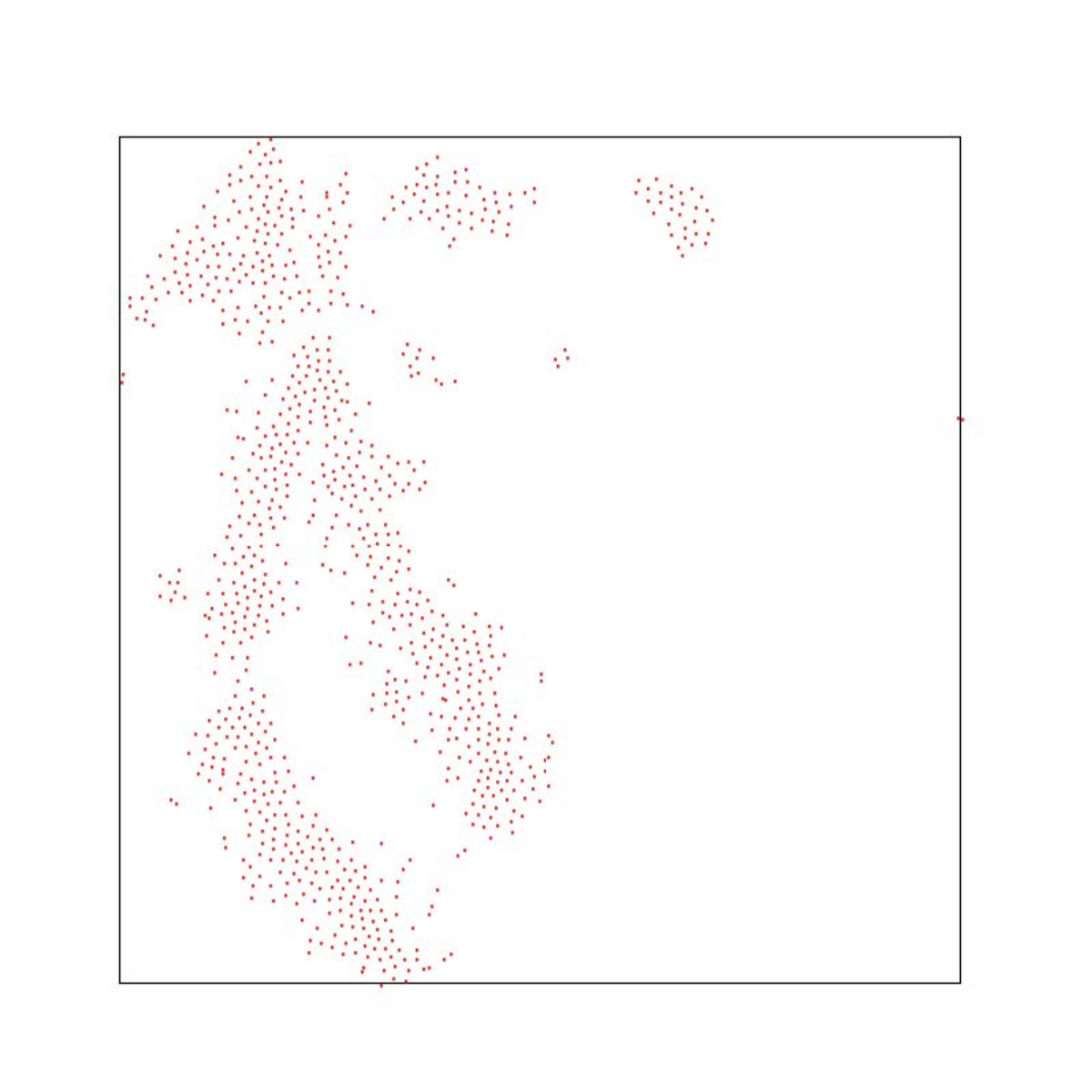}}
    \hfil 
    \subfloat[]{\includegraphics[height=1.25in]{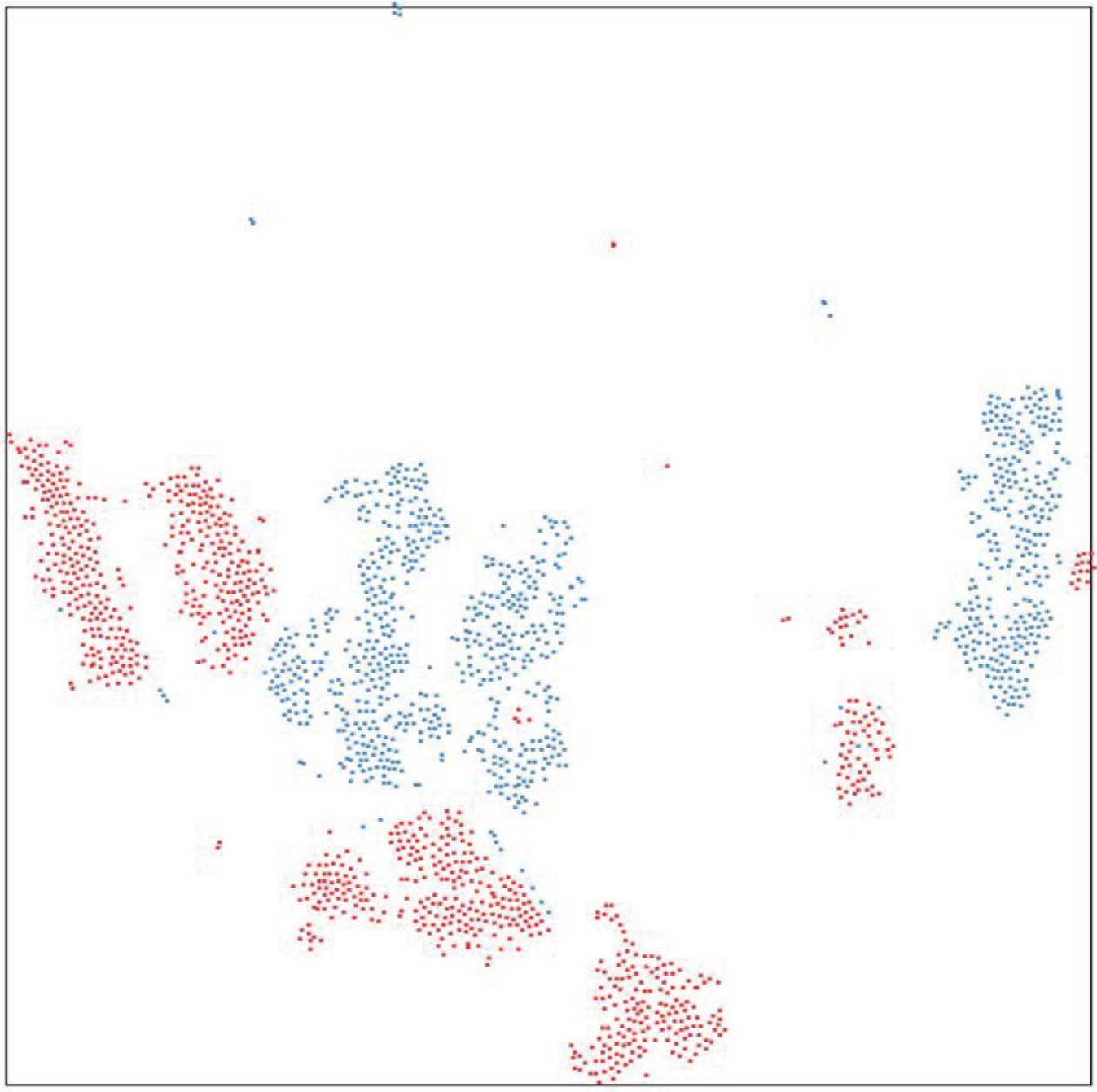}}
    \hfil
    \subfloat[]{\includegraphics[height=1.25in]{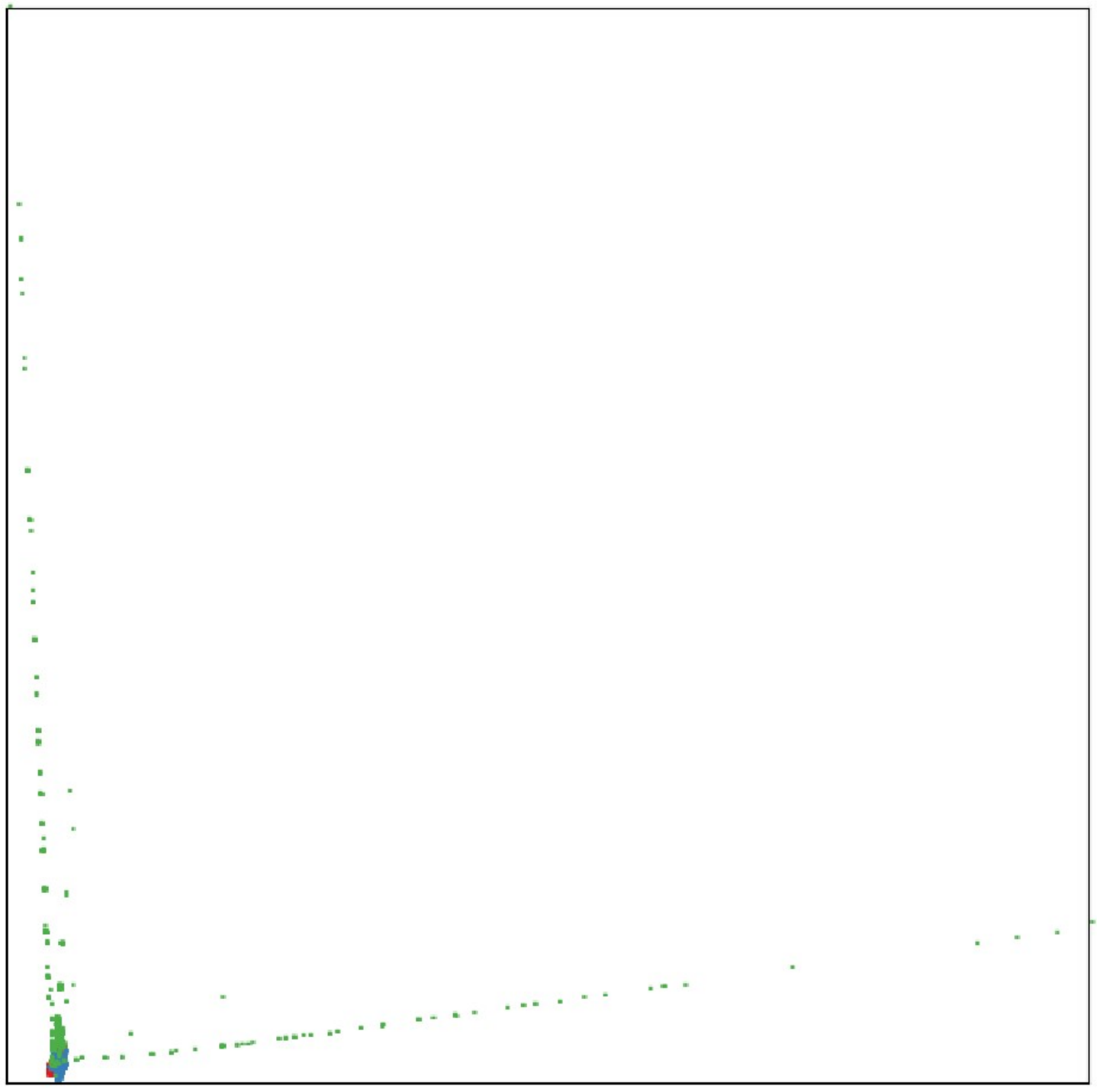}}
    \hfil
    \subfloat[]{\includegraphics[height=1.25in]{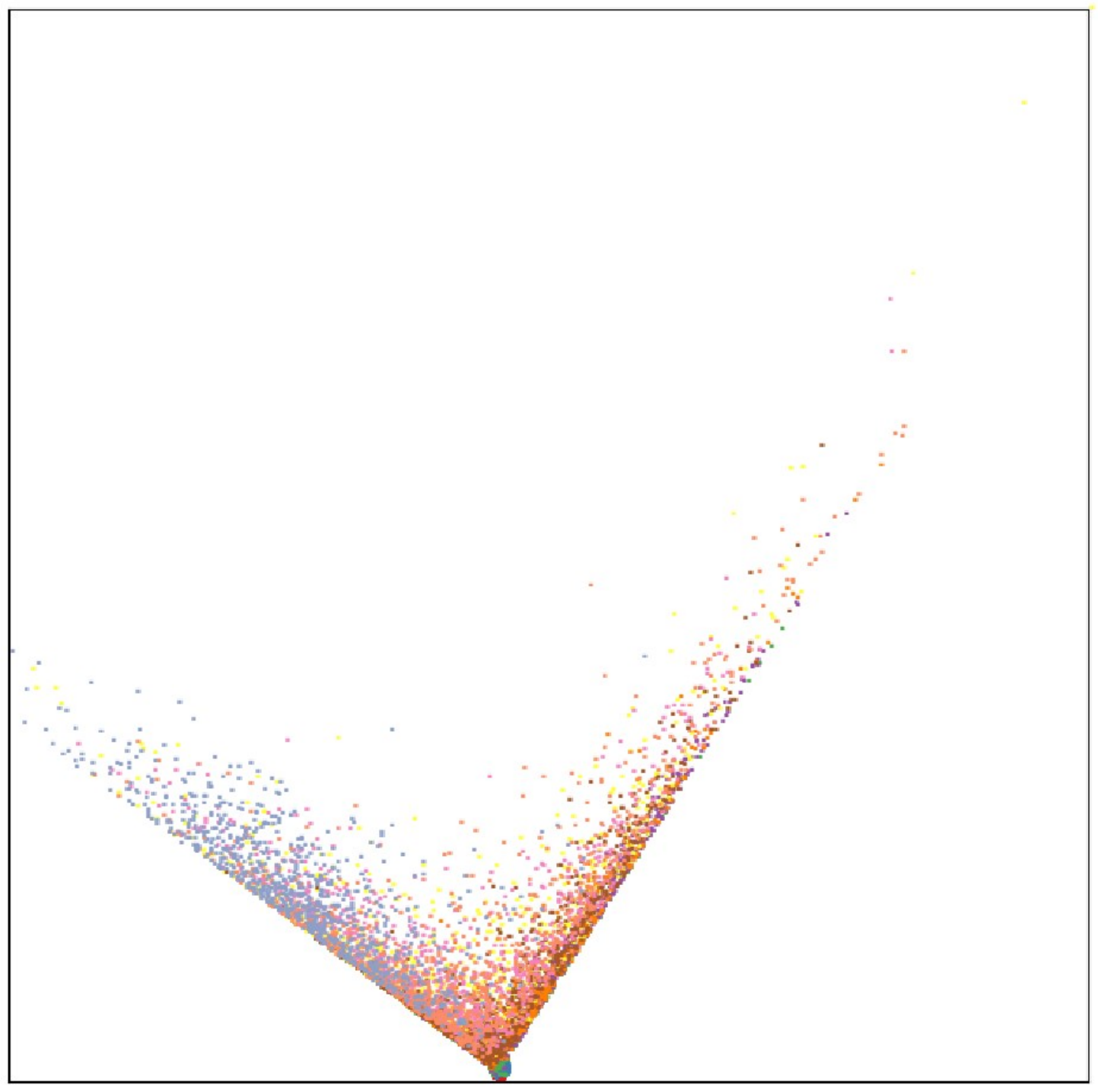}}
    \hfil
    \subfloat[]{\includegraphics[height=1.25in]{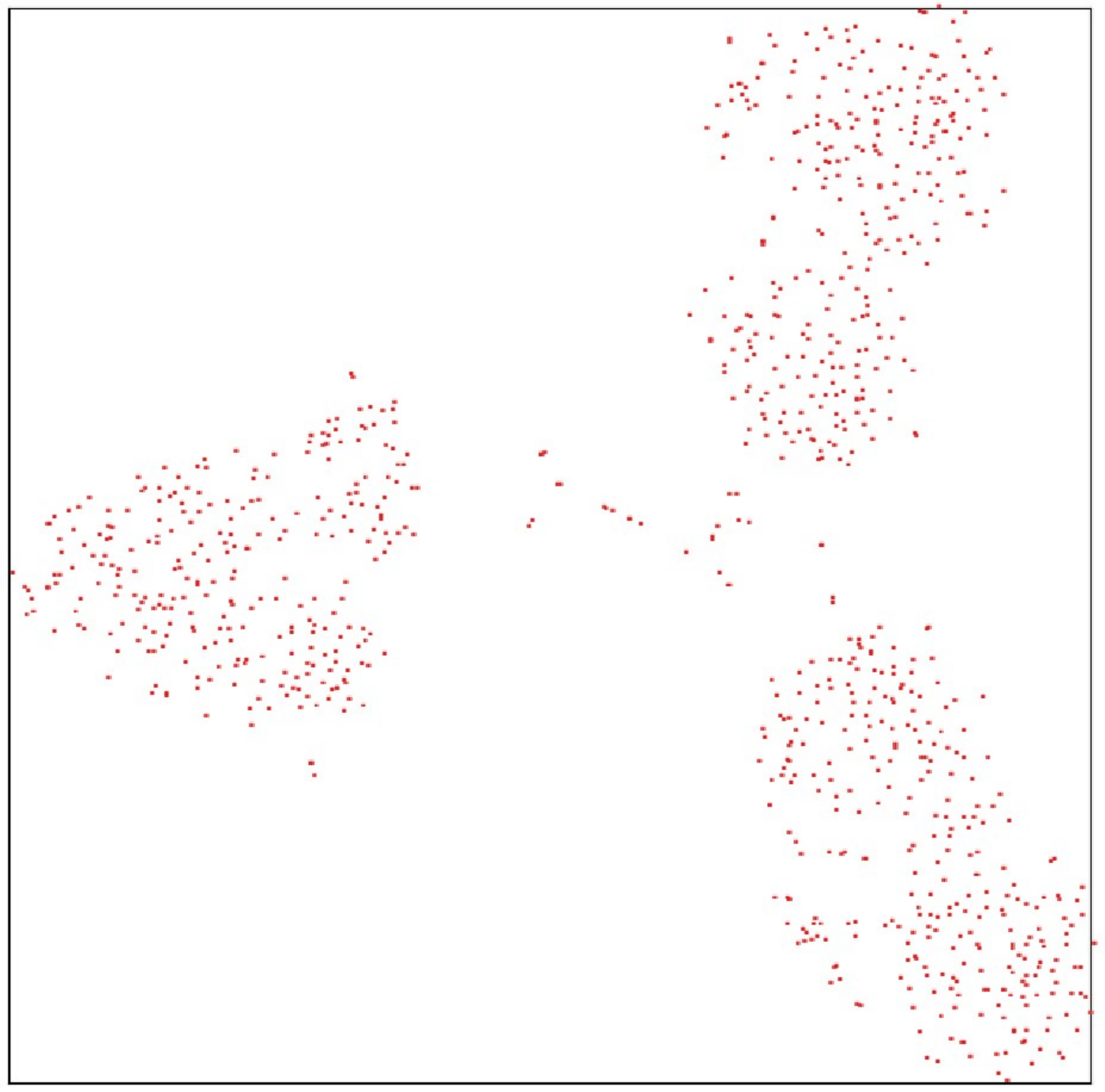}}
    \hfil
    \subfloat[]{\includegraphics[height=1.25in]{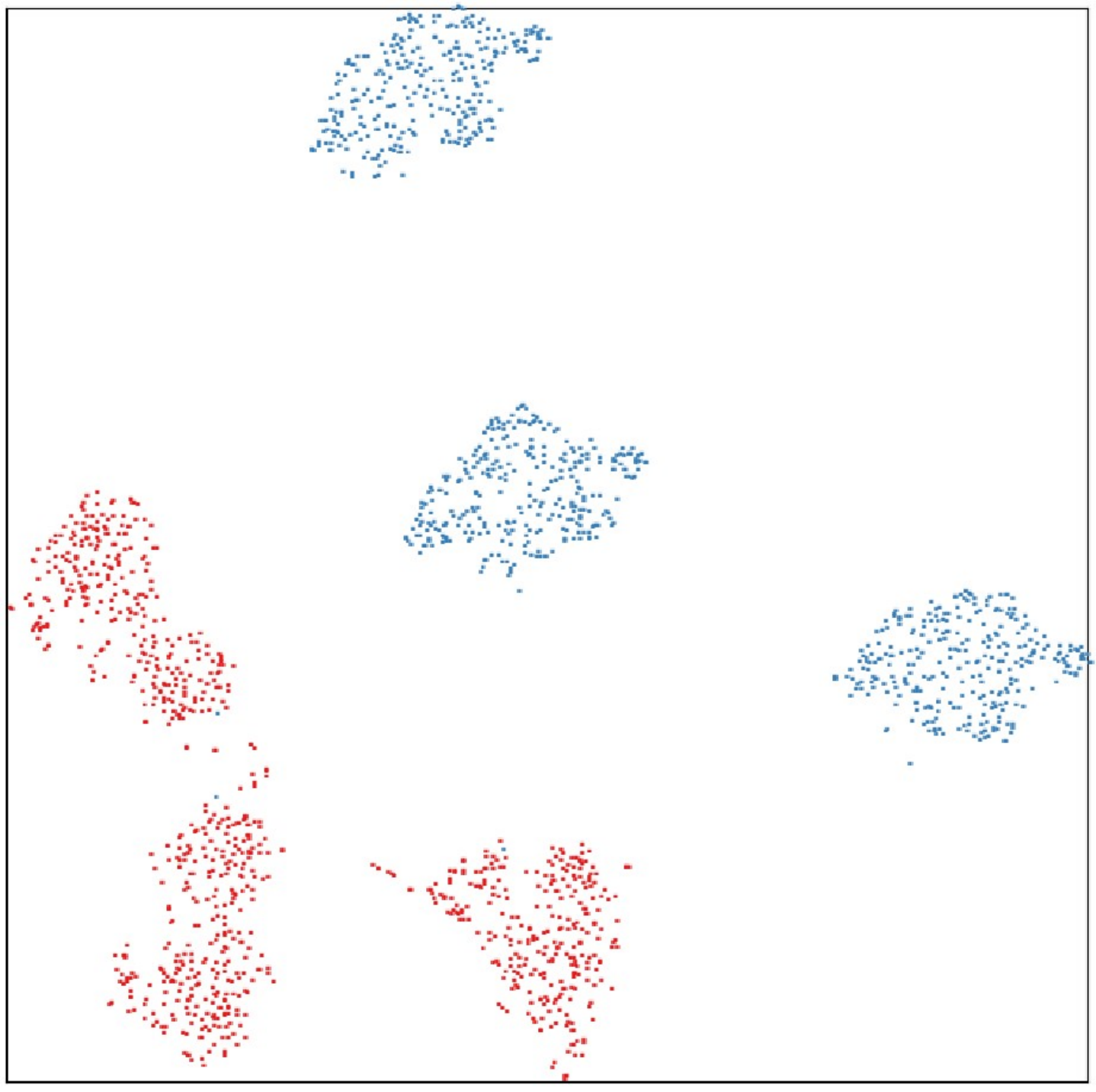}}
    \hfil
    \subfloat[]{\includegraphics[height=1.25in]{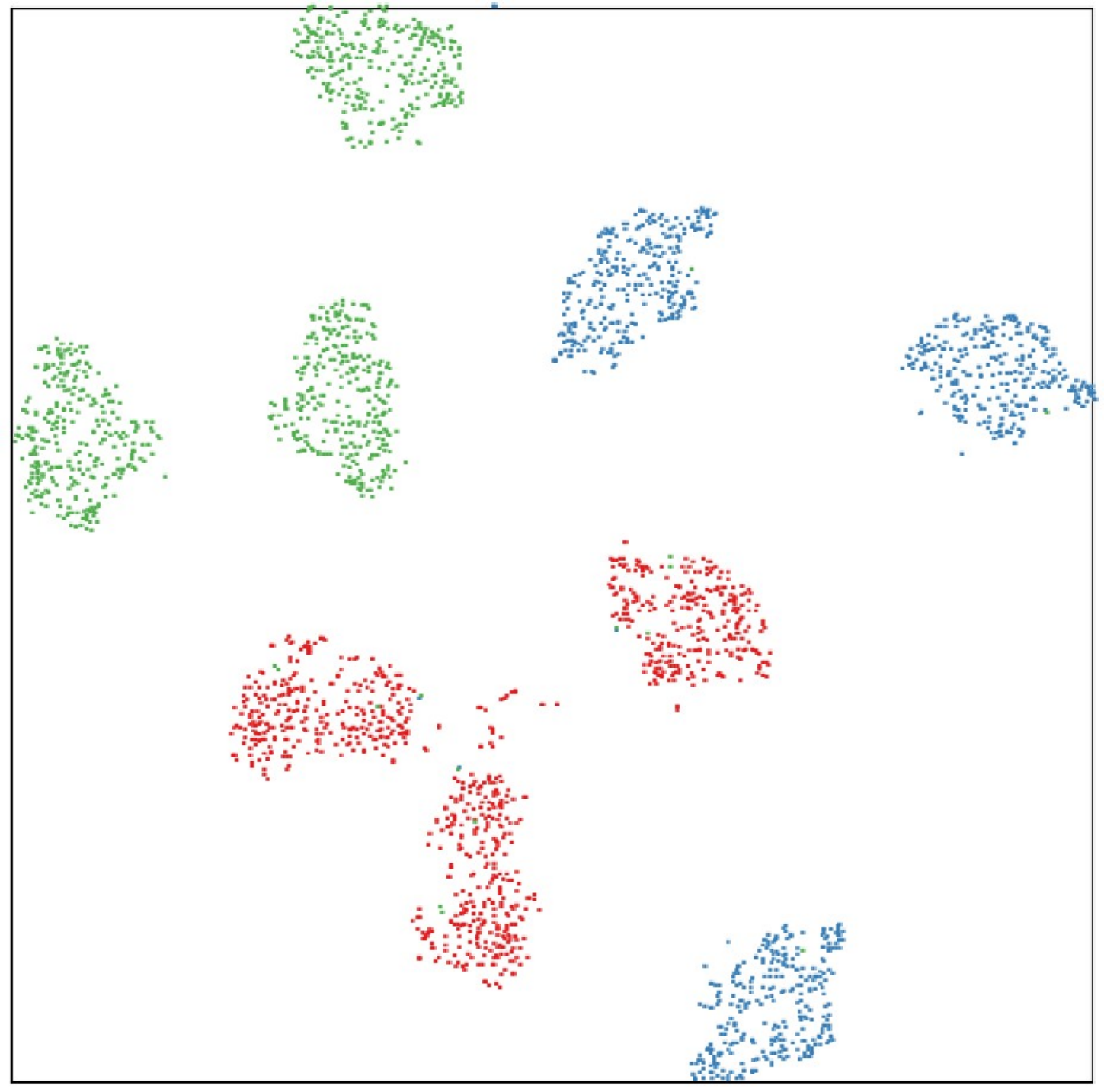}}
    \hfil
    \subfloat[]{\includegraphics[height=1.25in]{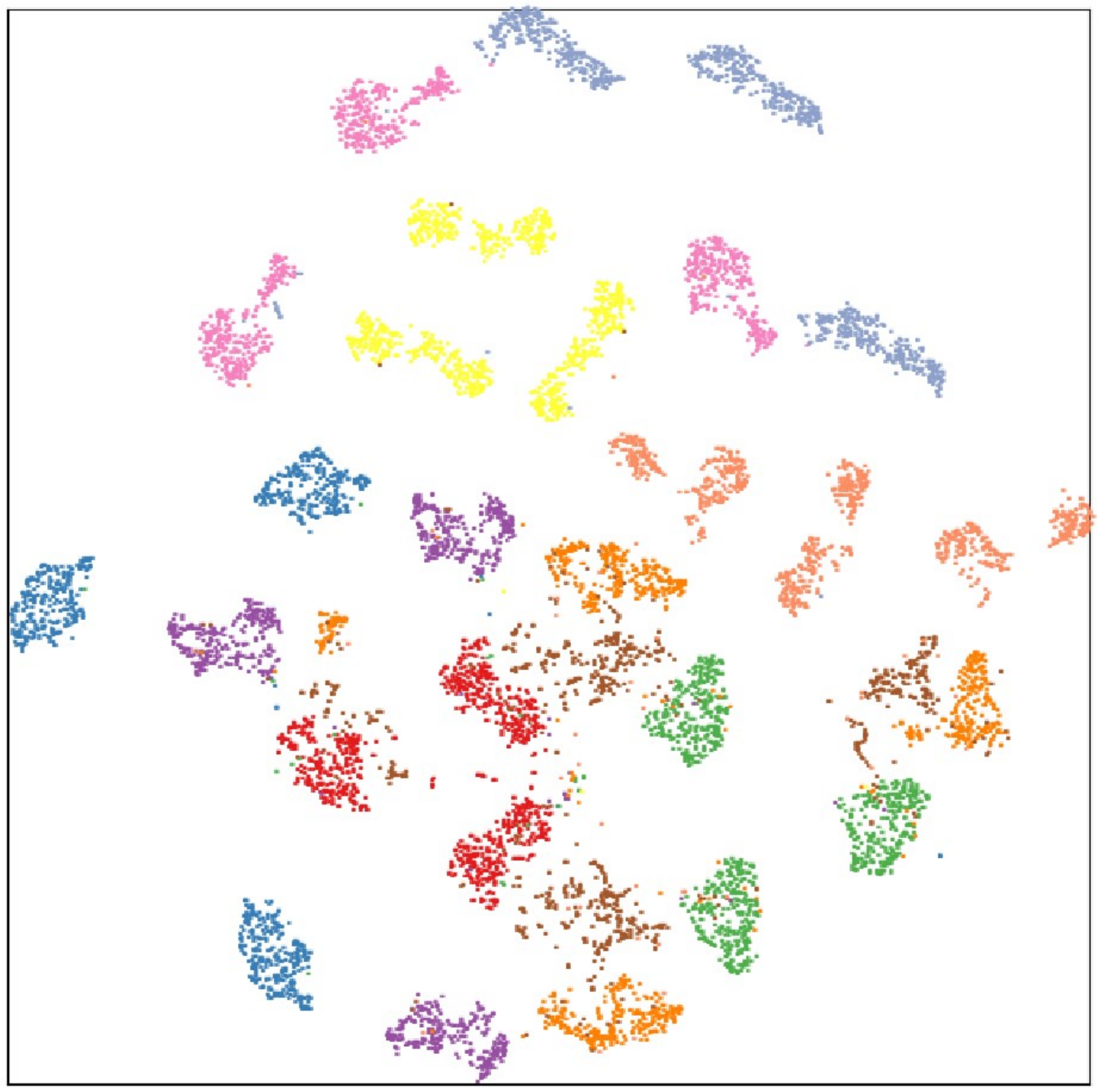}}
    \hfil
    \subfloat[]{\includegraphics[height=1.25in]{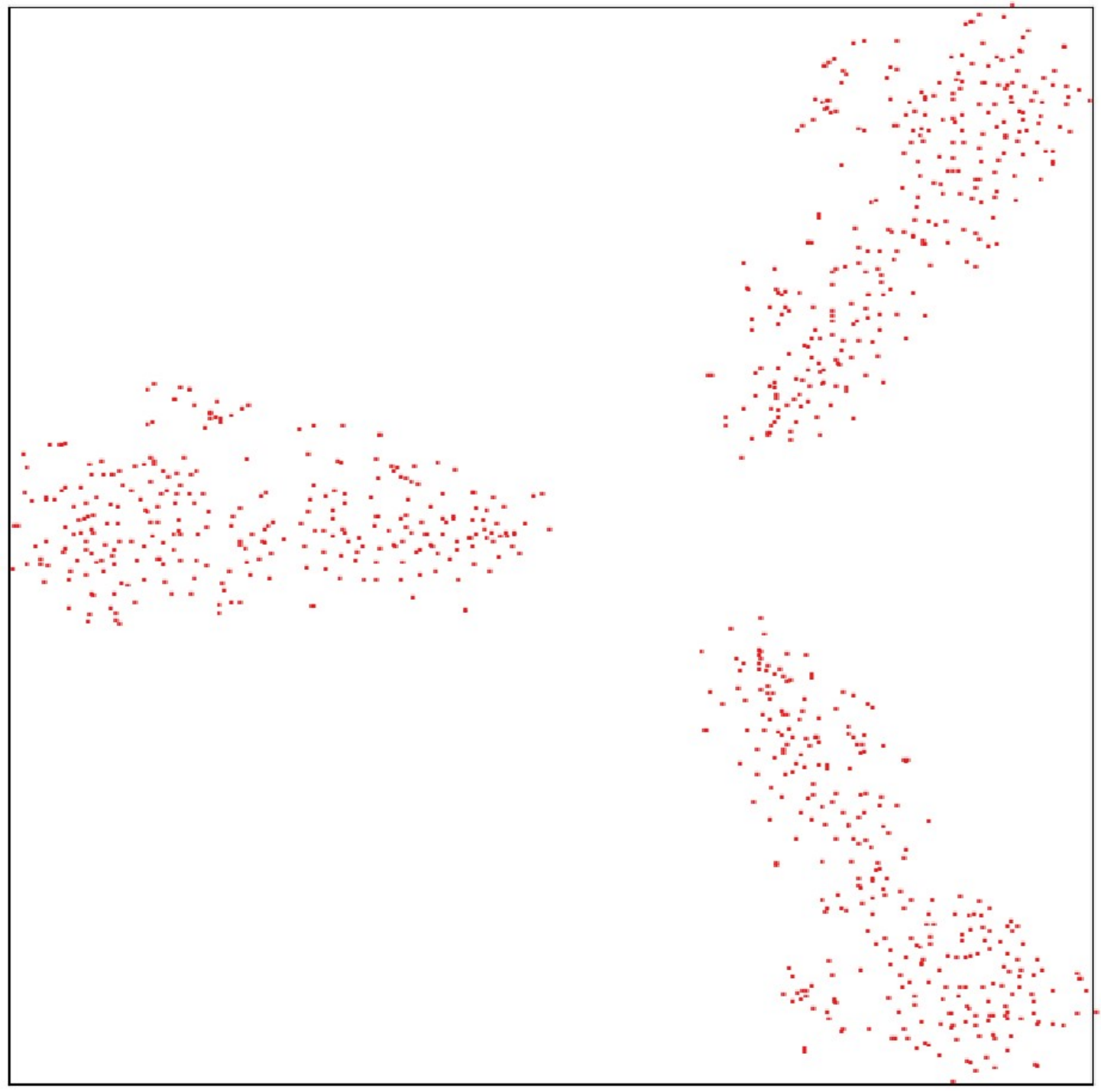}}
    \hfil
    \subfloat[]{\includegraphics[height=1.25in]{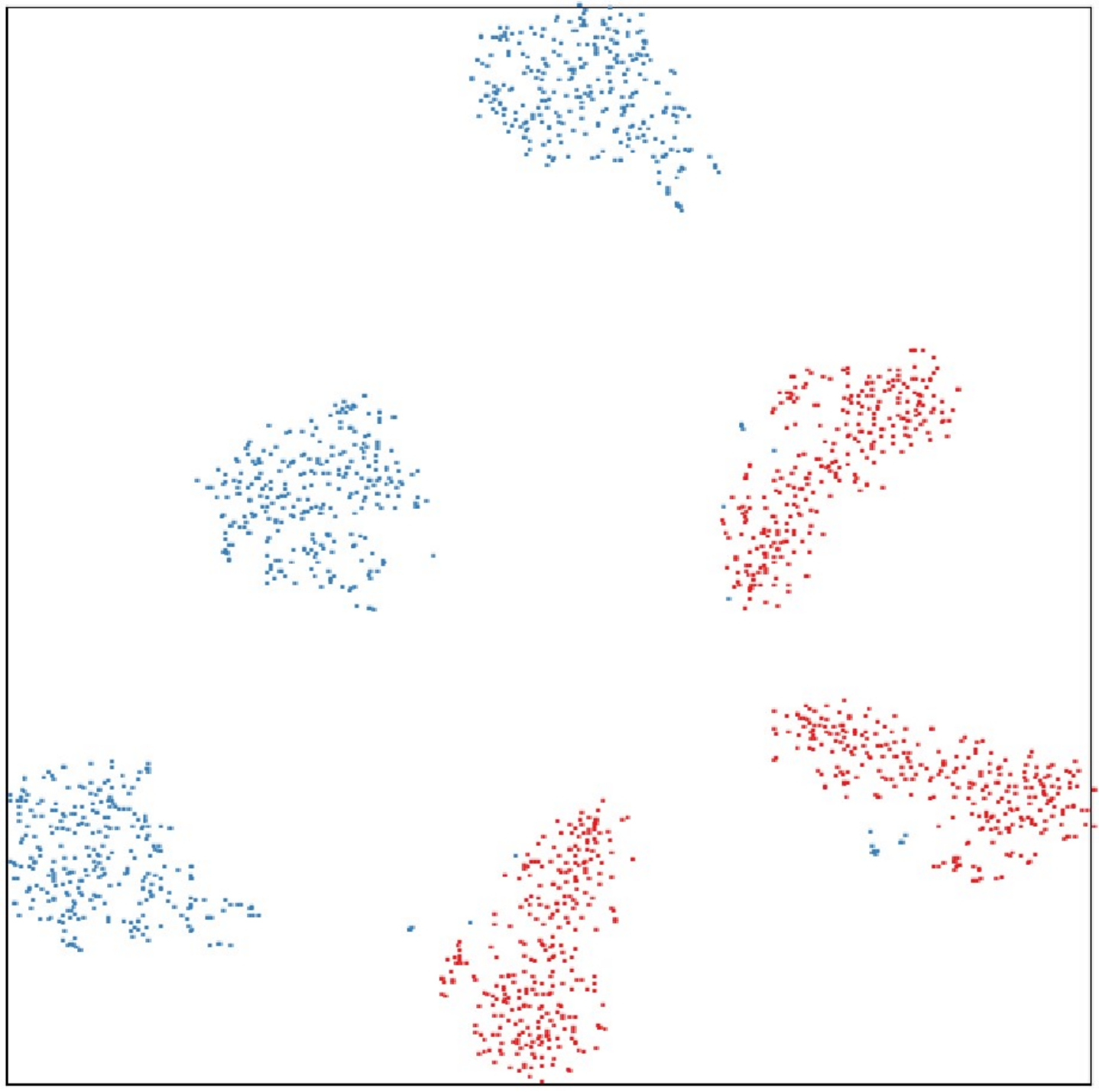}}
    \hfil
    \subfloat[]{\includegraphics[height=1.25in]{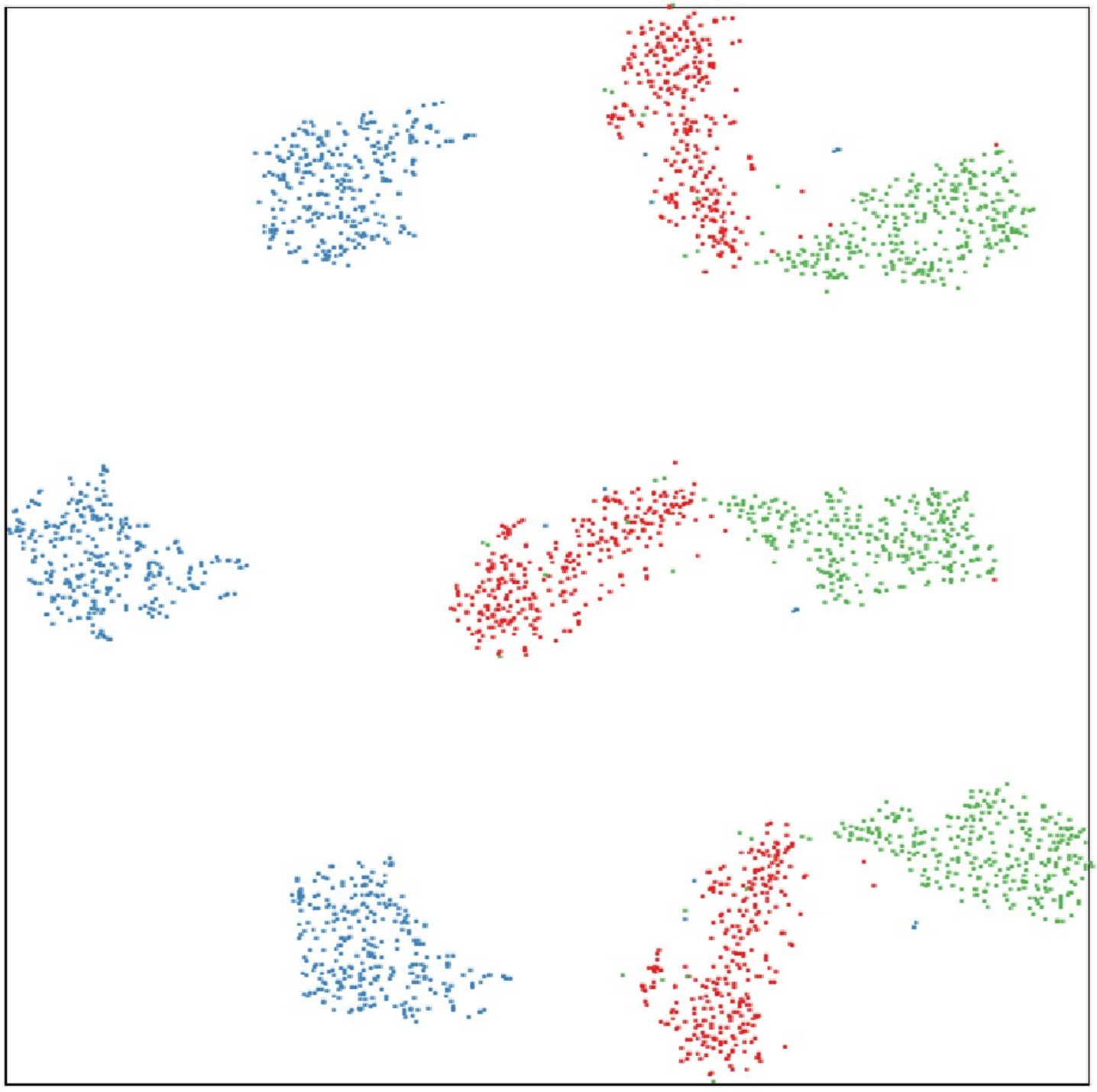}}
    \hfil
    \subfloat[]{\includegraphics[height=1.25in]{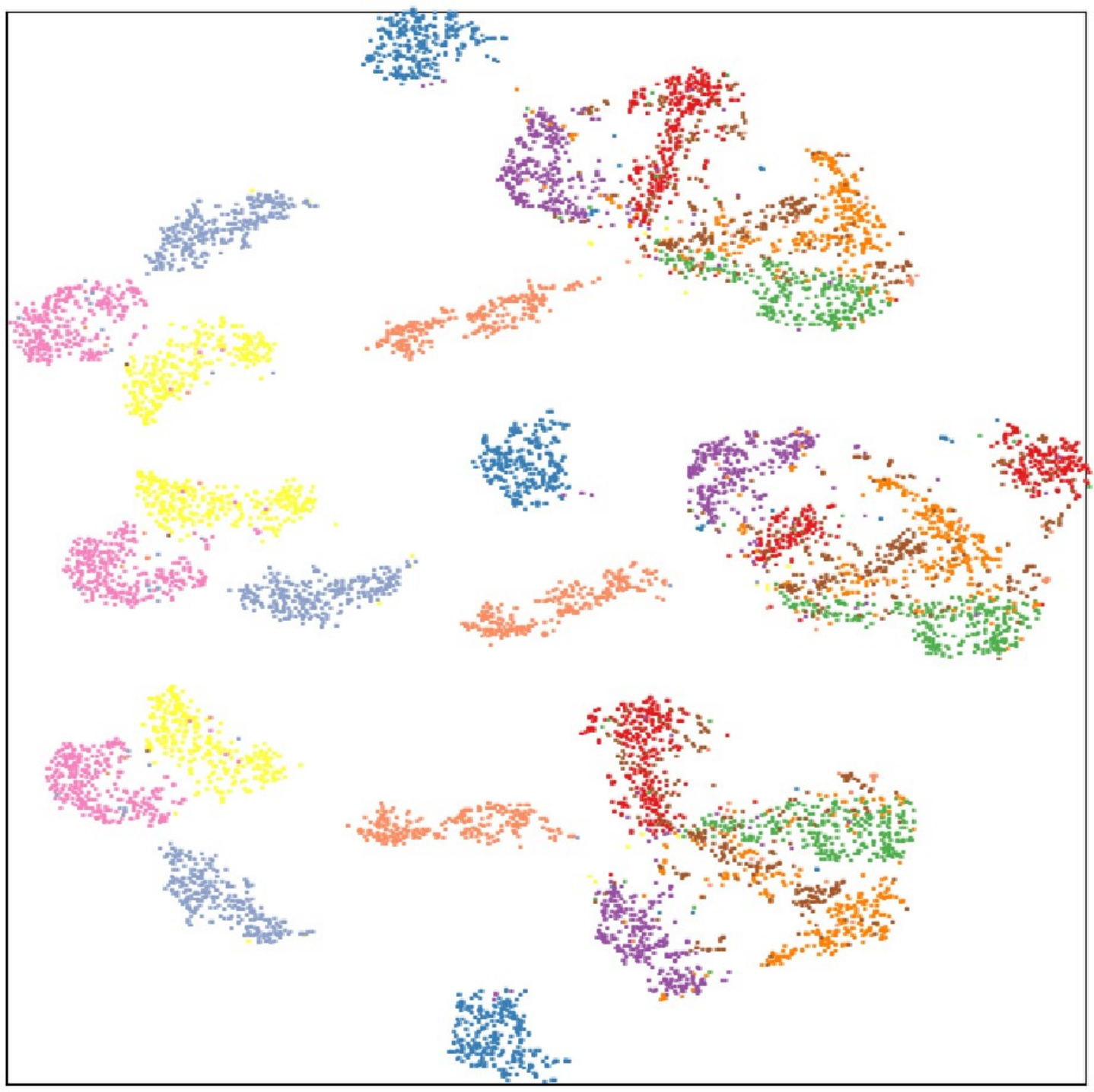}}
    \hfil
    \subfloat[]{\includegraphics[height=1.25in]{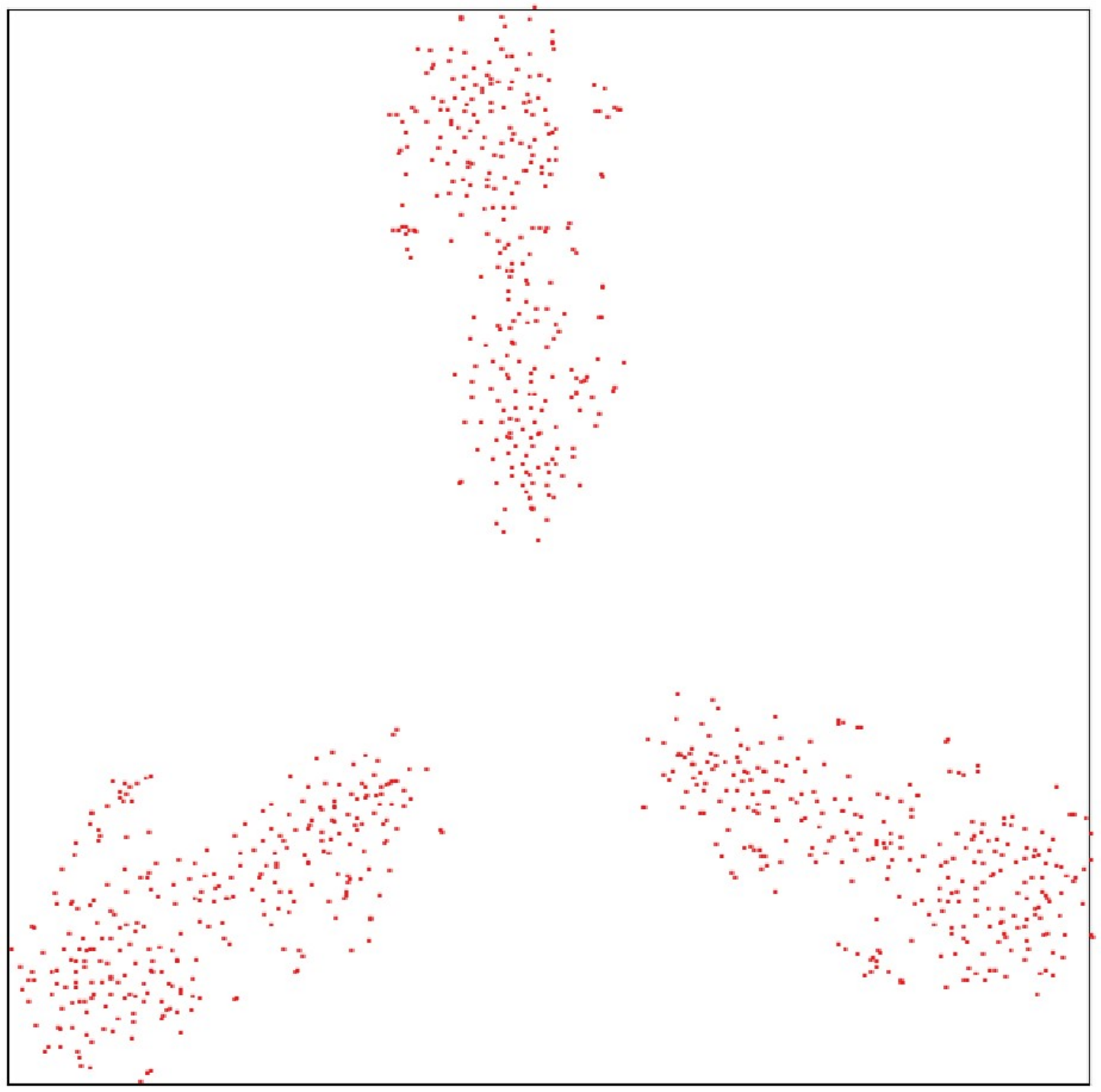}}
    \hfil
    \subfloat[]{\includegraphics[height=1.25in]{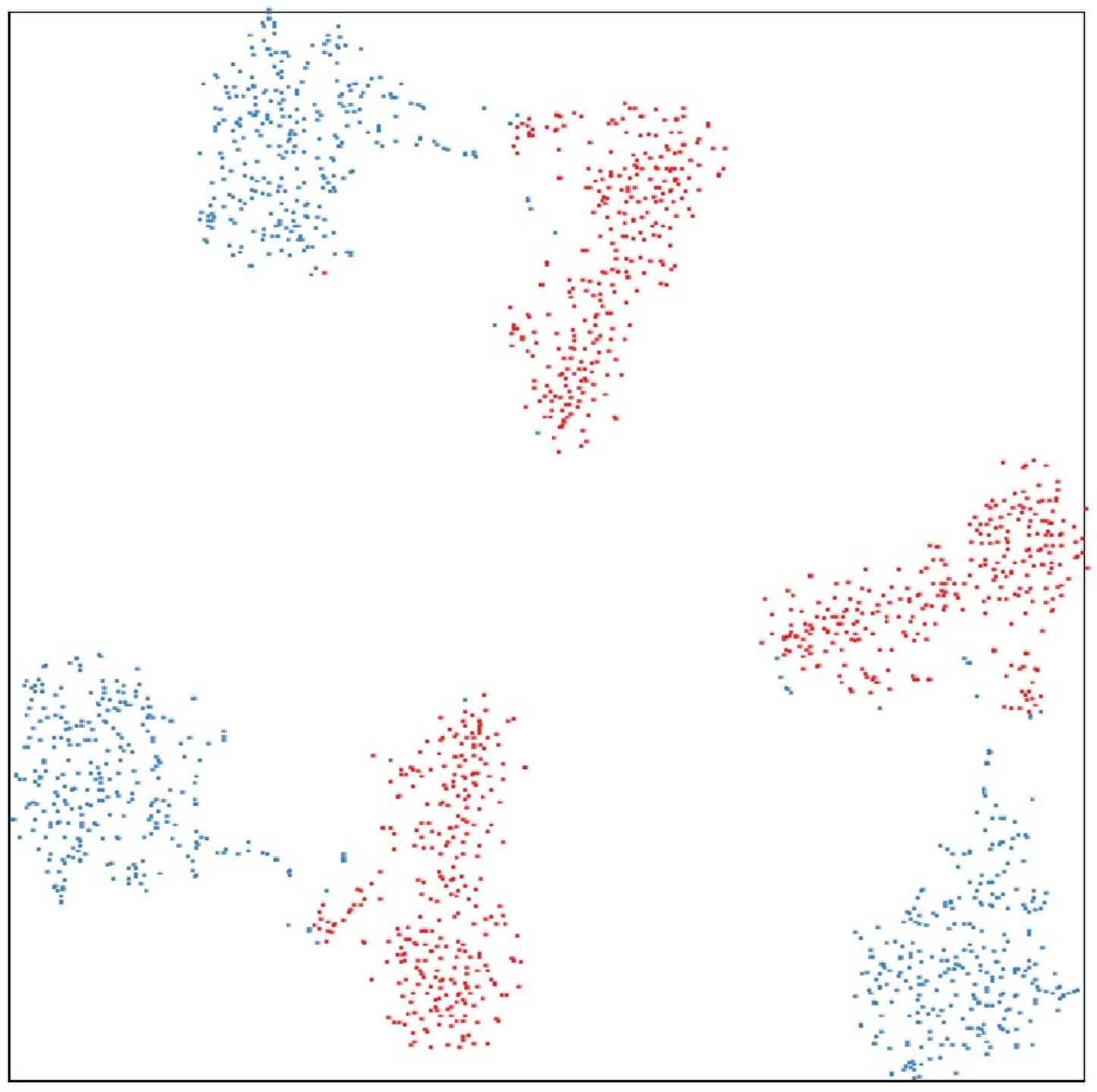}}
    \hfil
    \subfloat[]{\includegraphics[height=1.25in]{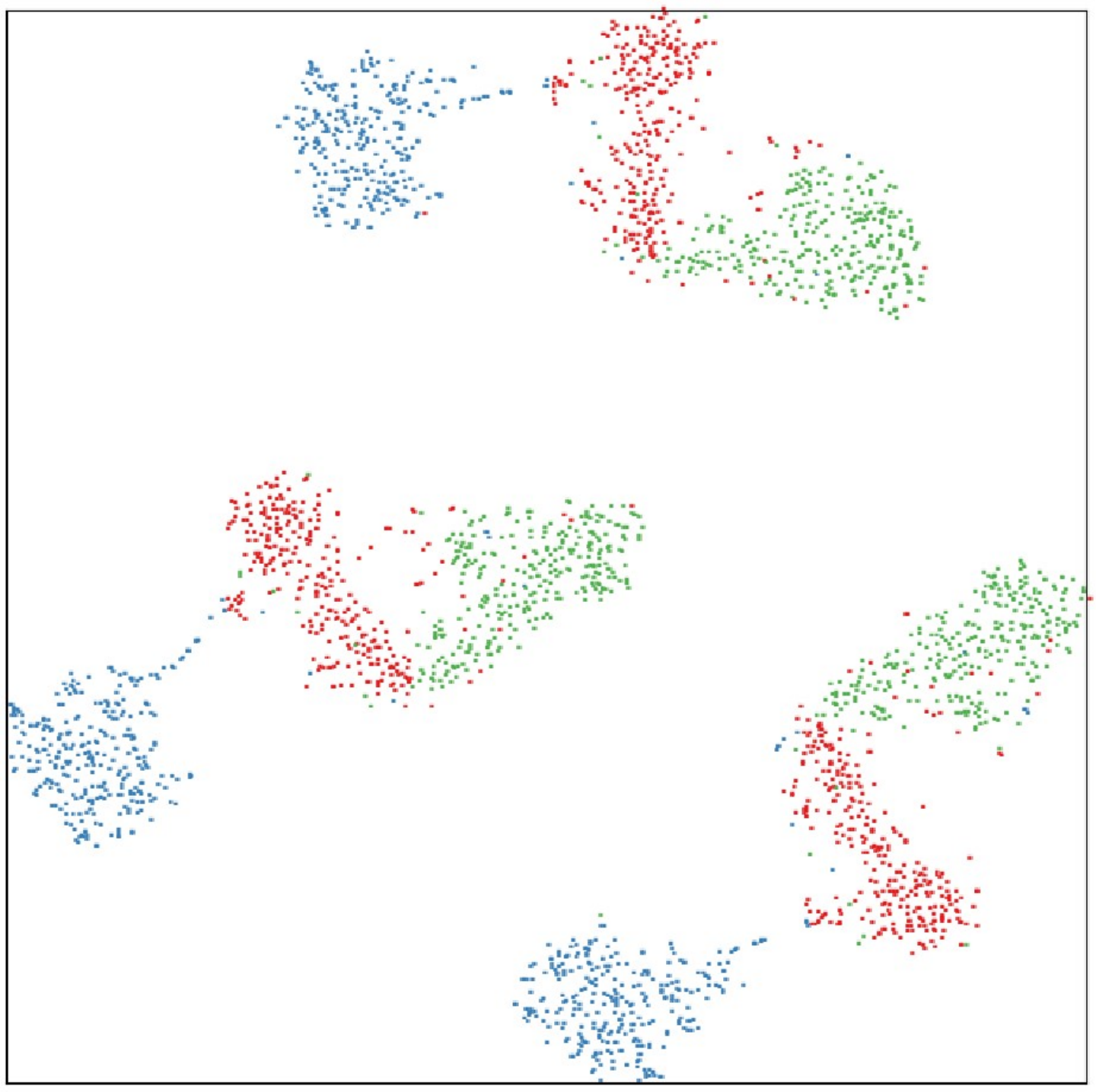}}
    \hfil
    \subfloat[]{\includegraphics[height=1.25in]{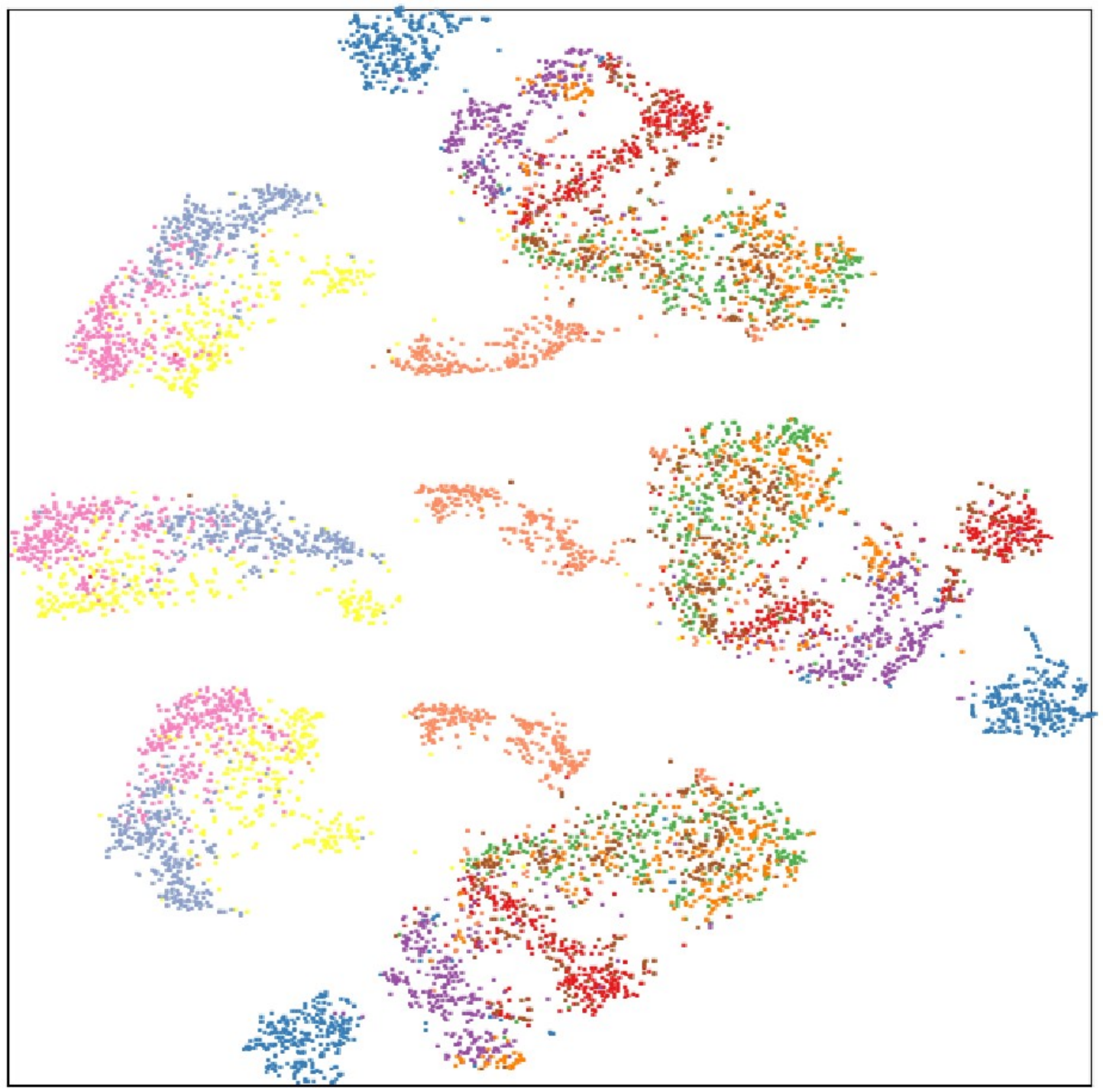}}
\caption{t-SNE visualization of our method under different levels of $\alpha$ value. Different colors represent different classes, which displays the incrementally fused continual views belonging to the first three classes and all. (a)-(d) $\alpha = 10^{-1}$; (e)-(h) $\alpha = 10^{0}$; (i)-(l) $\alpha = 10^{1}$; (m)-(p) $\alpha = 10^{2}$.}
\label{Visualization}
\end{figure}

Fig. \ref{Visualization} visualizes the outputs of the orthogonality fusion with t-SEN \cite{van2008visualizing} on the PFMNIST-10(3) dataset, in which a model needs to incrementally fuse a continual stream of 30 views for recognizing 10 classes. Given the space limitation, we only display the learned representations belonging to the first three classes and all. Note that the same color represents the same class and each class owns three views. In the case of $\alpha = 10^{-1}$, as shown in Fig. \ref{Visualization} (a)-(d), the model fuses the early views well but does not work as the number of views increases. In the case of $\alpha = 10^{0}$, the learned representation is more compact within the same classes and the inter-class boundaries are relatively clear. This implies the model can effectively fuse the complementary and consensus information among multi-view classes. In the cases of both $\alpha = 10^{1}$ and $\alpha = 10^{2}$, the model attends to concentrate on newly collected views but fails to retain the knowledge of past views.

\subsection{Ablation study}
Another set of experiments is performed to demonstrate the effectiveness of combining inter-view orthogonality fusion and inter-class selective weight consolidation, compared with that of giving equal treatment to each view or class. 
\textcolor{Blue}{To this end, besides EWC, we disentangle two additional networks that are built by either selective weight consolidation (SWC-Net) only or orthogonality fusion (OF-Net) only. Table \ref{Table_6} reports the results of different components or combinations. We observe that EWC and SWC-Net suffer from serious performance degradation since they regard each view as a separate task. This not only neglects the fusion of the complementary and consensus information but also exacerbates the challenges existed in MVCIL;} And OF-Net amounts to applying two layer-wise orthogonality fusions, with one for views and another for classes. This sometimes requires sufficient memory to maintain these orthogonal subspaces and is more vulnerable to increasing the generalization bound error \cite{bartlett2002rademacher} during sequential training. By contrast, the proposed method outperforms the competitors on the four different datasets, demonstrating its effectiveness in incrementally recognizing new classes from a continual stream of views.

\begin{table}[tbp]
\caption{\textcolor{Blue}{The effectiveness of combining inter-view orthogonality fusion and inter-class selective weight consolidation used in our method. We report the results on four different datasets, measured by the Avg Acc.}}
\label{Table_6}
\centering
\begin{tabular}{lcccc}
\toprule
Component & COIL-20(4) & COIL-20(8) & AwA-50(2) & PIE-68(3)  \\ \midrule
EWC &22.55$\pm$8.24 &27.02$\pm$7.24 &2.10$\pm$0.05 &1.48$\pm$0.02  \\
SWC-Net &23.06$\pm$7.10 &26.18$\pm$6.55 &2.16$\pm$0.02 &1.49$\pm$0.00  \\
OF-Net &80.44$\pm$3.46 &83.59$\pm$2.24 &58.03$\pm$0.51 &53.51$\pm$1.02  \\ \midrule
Ours &\bf 85.14$\pm$2.20 &\bf 86.73$\pm$1.48 &\bf 63.86$\pm$0.53 & \bf62.34$\pm$1.02 \\ \bottomrule  
\end{tabular}
\end{table}

\subsection{Investigating the generalization of MVCNet to familiar views}

\begin{table}[tbp]
\setlength{\tabcolsep}{1.2mm} 
\caption{Experimental results on the non-overlapping views of each class that have not been used for training. ‘Acc$\downarrow$’ denotes the performance gap compared to the counterparts of fully shared views between training and test.}
\label{Table_7}
\centering
\begin{tabular}{lccc|ccc}
\toprule
\multirow{2}{*}{Method} & \multicolumn{3}{c}{COIL-20(4)} & \multicolumn{3}{c}{COIL-20(8)} \\ \cmidrule{2-4} \cmidrule{5-7}
 & Offline Acc & Avg Acc & Acc$\downarrow$ & Offline Acc & Avg Acc & Acc$\downarrow$ \\ \midrule
deepCCA &62.73$\pm$0.40  & - &-6.16  &67.68$\pm$0.43  & - &-28.99  \\
InfoMax &64.33$\pm$3.34  & - &-6.34  &88.78$\pm$3.27  & - &-4.76  \\
MIB     &65.82$\pm$1.63  & - &-6.55  &82.96$\pm$2.56  & - &-11.54  \\
MV-InfoMax &66.95$\pm$2.83  & - &-9.44  &\bf 83.22$\pm$2.17  & - &-12.88  \\
VAE     &63.05$\pm$1.54  & - &-23.61  &84.07$\pm$1.79  & - &-8.43  \\ 
CPM-Net &69.28$\pm$0.50  & - &-17.37  &73.70$\pm$0.84  & - &-22.84  \\
TMC     &\bf 78.80$\pm$0.43  & - &-16.87  &83.14$\pm$0.85  & - &-14.09  \\ \midrule
EWC  & - &24.06$\pm$9.76  &-1.56  & - &22.25$\pm$5.25  &-4.83  \\
PCL  & - &55.11$\pm$4.14  &-2.29  & - &55.23$\pm$2.03  &-3.31  \\
FS-DGPM & - &69.76$\pm$3.49  &-2.18  & - &66.95$\pm$3.21  &-2.85  \\
OWM  & - &70.94$\pm$5.49  &-4.39  & - &74.83$\pm$2.44  &-3.38  \\
GEM  & - &73.24$\pm$1.25  &-2.09  & - &78.52$\pm$4.16  &-4.29  \\
LOGD & - &75.42$\pm$4.10  &-2.73  & - &78.73$\pm$2.78  &-1.42  \\
IL2M & - &80.86$\pm$2.97  &-1.51  & - &82.22$\pm$2.94  &-2.47  \\ \midrule
Ours & - &\bf 84.58$\pm$3.35  &\bf -0.98  & - &\bf 85.46$\pm$3.01  &\bf -1.30  \\ \bottomrule     
\end{tabular}
\end{table}

Before concluding our work, we place the proposed method in a more general MVCIL scenario where partial views belonging to a certain class are excluded from the training process but used for testing. Specifically, there is no overlap between the views of the same class during both the training and testing stages. This investigates the generalization of a model to familiar but untrained views during inference time. Take COIL-20(4) and COIL-20(8) datasets as examples. Note that they are drawn by step sizes of $90^{\circ}$ and $45^{\circ}$ from 20 classes in which each class is represented by 4 views and 8 views, respectively. Here we select 3 views of each class from the COIL-20(4) dataset for training a model and the remaining one for testing its generalization. Similarly, we train a model using 7 views of each class from the COIL-20(8) dataset and the remaining one for testing.

Table \ref{Table_7} first records the results of different methods on these familiar views and then shows the performance gap compared to the counterparts of fully shared views between training and test, as reported in Tables \ref{Table_2} and \ref{Table_3}. Note that, in these two cases, MVL baselines only remember the class learned last time and the results on Avg Acc are similar. Hence, we omit them and use the Offline Acc instead. The results presented in Table \ref{Table_7} demonstrate that all methods exhibit a certain level of performance degradation when confronted with familiar views. On the whole, MVL baselines exhibit notable vulnerability even in the case of being trained offline over all classes; CIL baselines struggle in the MVCIL paradigm. By contrast, the proposed method is significantly superior to them. This indicates the robust generalization of MVCNet to familiar views.

\section{Conclusion}
\label{Sec_Con}
Existing multi-view learning methods typically assume that all views can be simultaneously accessed and a model only works on a prepared multi-view dataset with fixed classes, which is less practical in an open-ended environment. To address the limitations, this paper investigates a novel multi-view class incremental learning paradigm by drawing inspiration from recent advances in incremental learning. The proposed method called MVCNet offers an attractive way of incrementally classifying new classes from a continual stream of views, requiring no access to earlier views of data. To the best of our knowledge, this is yet underexplored in the multi-view learning community. The sufficient experiments demonstrate that our method is applicable to the multi-view class incremental learning scenario. In future research, we intend to extend this idea to encompass cases involving incomplete views and trusted views, i.e., we learn an incremental classifier from a realistic scenario with more abundant view-level characteristics.

\section*{CRediT authorship contribution statement}
\textbf{Depeng Li:} Conceptualization, Methodology, Software, Roles/Writing – original draft, Writing – review \& editing. \textbf{Tianqi Wang:} Conceptualization, Software, Investigation, Visualization. \textbf{Junwei Chen:} Software, Investigation, Validation. \textbf{Kenji Kawaguchi:} Writing - review \& editing, Supervision, Resources. \textbf{Cheng Lian:} Writing - review \& editing. \textbf{Zhigang Zeng:} Writing - review \& editing, Supervision, Funding acquisition.

\section*{Acknowledgements}
This work was supported by the National Key R\&D Program of China under Grant 2021ZD02-01300, the Fundamental Research Funds for the Central Universities under Grant YCJJ202203012, the State Scholarship Fund of China Scholarship Council under Grant 202206160045, the National Natural Science Foundation of China under Grants U1913602 and 61936004, the Innovation Group Project of the National Natural Science Foundation of China under Grant 61821003, and the 111 Project on Computational Intelligence and Intelligent Control under Grant B18024.





\bibliographystyle{elsarticle-num}
\bibliography{mybibfile}






\end{document}